\documentclass[acmsmall]{acmart}
\usepackage{multirow}
\usepackage{makecell}
\usepackage{enumitem}
\setlist{leftmargin=3.5mm}

\usepackage{tabularx}
\usepackage{xcolor}
\usepackage{url}
\usepackage{xspace}
\usepackage{caption}
\usepackage{subcaption}
\usepackage{float}
\usepackage{hyperref}
\usepackage{cleveref}
\usepackage{float}
\hypersetup{
    colorlinks=true,
    linkcolor=blue,
    filecolor=magenta,      
    urlcolor=cyan,
}
\definecolor{Author1}{HTML}{D03416}

\AtBeginDocument{%
  \providecommand\BibTeX{{%
    \normalfont B\kern-0.5em{\scshape i\kern-0.25em b}\kern-0.8em\TeX}}}

\setcopyright{acmcopyright}
\copyrightyear{2022}
\acmYear{2022}
\acmDOI{XXXXXXX.XXXXXXX}

\begin{document}

\title{Going Beyond XAI: A Systematic Survey for Explanation-Guided Learning}
% \ray{let's add Beyond XAI?}

\author{Yuyang Gao}
\email{yuyang.gao@emory.edu}
\affiliation{%
  \institution{Emory University}
  \city{Atlanta}
  \state{Georgia}
  \country{USA}
  \postcode{30322}
}
\author{Siyi Gu}
\email{carrie.gu@emory.edu}
\affiliation{%
 \institution{Emory University}
 \city{Atlanta}
 \state{GA}
 \country{USA}
}
\author{Junji Jiang}
\email{jjian50@emory.edu}
\affiliation{%
 \institution{Emory University}
 \city{Atlanta}
 \state{GA}
 \country{USA}
}
\author{Sungsoo Ray Hong}
\email{shong31@gmu.edu}
\affiliation{%
 \institution{George Mason University}
 \city{Fairfax}
 \state{VA}
 \country{USA}
}
\author{Dazhou Yu}
\email{dazhou.yu@emory.edu}
\affiliation{%
 \institution{Emory University}
 \city{Atlanta}
 \state{GA}
 \country{USA}
}
\author{Liang Zhao}
\authornote{Corresponding author}
\email{liang.zhao@emory.edu}
\affiliation{%
 \institution{Emory University}
 \city{Atlanta}
 \state{GA}
 \country{USA}
}

\renewcommand{\shortauthors}{Gao et al.}

\begin{abstract}
    As the societal impact of Deep Neural Networks (DNNs) grows, the goals for advancing DNNs become more complex and diverse, ranging from improving a conventional model accuracy metric to infusing advanced human virtues such as fairness, accountability, transparency (FaccT), and unbiasedness.
    Recently, techniques in Explainable Artificial Intelligence (XAI) are attracting considerable attention, and have tremendously helped Machine Learning (ML) engineers in understanding AI models. 
    However, at the same time, we started to witness the emerging need beyond XAI among AI communities; based on the insights learned from XAI, how can we better empower ML engineers in steering their DNNs so that the model's reasonableness and performance can be improved as intended?
    This article provides a timely and extensive literature overview of the field \emph{\textbf{Explanation-Guided Learning}} (EGL), a domain of techniques that steer the DNNs' reasoning process by adding regularization, supervision, or intervention on model explanations.
    In doing so, we first provide a formal definition of EGL and its general learning paradigm.
    Secondly, an overview of the key factors for EGL evaluation, as well as summarization and categorization of existing evaluation procedures and metrics for EGL are provided.
    Finally, the current and potential future application areas and directions of EGL are discussed, and an extensive experimental study is presented aiming at providing comprehensive comparative studies among existing EGL models in various popular application domains, such as Computer Vision (CV) and Natural Language Processing (NLP) domains.
\end{abstract}




\maketitle
\section{Introduction}

% \textbf{1. Why model explanation is important? Why only have the explanation is not enough}
In recent years, techniques in Explainable Artificial Intelligence (XAI) are attracting considerable attention~\cite{adadi2018peeking, arrieta2020explainable, guidotti2018survey}, and have gradually become the dominating ways that connect the way Deep Neural Networks (DNNs) work and human reasoning~\cite{hong2020human, li2021survey}. 
As DNNs cannot provide human sensible ``global structure'' of how the model works unlike white-box models, XAI has become an imperative tool that Machine Learning (ML) engineers always use to ``make sense'' of the way their models work~\cite{gao2022aligning}.
In recent years, many XAI techniques have been proposed in an effort to open the ``black box'' of DNNs~\cite{guidotti2018survey}, such as techniques that provide saliency maps for understanding which sub-parts (i.e., features) in an instance are most responsible for the model prediction~\cite{zhou2016learning, selvaraju2017grad, montavon2019layer, bach2015pixel, montavon2017explaining}.
%\ray{citations dead here}.
Despite the recent fast progress on XAI techniques for DNNs, the majority of the research body in XAI put focus on handling ``how to generate the explanations'' while showing less attention to advanced questions like ``whether the explanations are reasonable/accurate'', ``what if the explanations are unreasonable/inaccurate'', and most importantly, ``how to adjust the model to generate more reasonable/accurate explanations in the future''.
We are starting to witness the emerging need beyond XAI; based on the insights learned from XAI, how can we better steer DNNs such that their future behavior can be improved from the insights learned from XAI techniques?
We argue that understanding how to convert insights learned from XAI-driven techniques to steer DNNs would be the key to realizing the DNNs to be more powerful, fair, accountable, transparent, unbiased, and trustworthy, unraveling many real-world application areas.
%Such capability is closely associated with our capability of 
%that can further advance the way we apply DNNs in real application areas 
%development at the verge of science.%in  where a powerful and trustworthy model is on demand. 

% \textbf{2. What existing works have been done after getting the explanation? why they are still not enough compared with EGL?}
In recent years, several new areas have emerged which aim at gaining a thorough grasp of the model behavior through the model explanation.
Explanatory Debugging is one area of research that has gained  popularity~\cite{kulesza2015principles, lertvittayakumjorn2021explanation, wexler2019if}. 
Interactive techniques and systems were developed to enable human users to interactively select features of interest and then investigate how the model behaves in the resulting subspaces for debugging purposes. 
Another interesting area of research compared the explanation provided by DNNs and the explanation provided by humans to gain a better understanding of the models' behavior~\cite{tan2021diversity, das2017human}.
Although the aforementioned studies are capable of providing more insights about whether the explanations are accurate or reasonable, they are yet to be sufficient for further handling how we can learn from those mistakes, and consequently adjust the model to get better quality explanations and enhance the model performance.

% \textbf{3. What is EGL, and what are the unique challenges?}
Recently, a new line of research that aims to intervene ML model's behavior through XAI techniques has started to emerge. 
In particular, the approaches jointly improve DNNs in terms of both their explainability and generalizability by applying additional supervision signals or prior knowledge onto the model reasoning process to direct the model explanation derived from established XAI techniques.
This direction is generally named Explanation Guided Learning (EGL)~\cite{56, 59, 68, 26}, while several other terms such as \textit{Explanation Supervision}~\cite{gao2021gnes, gao2022res, gao2022aligning}, \textit{Attention Supervision}~\cite{73, 74}, \textit{Explanation Alignment}~\cite{42, 67}, as well as \textit{Learning from Explanation}~\cite{10, 12, 61} are also frequently used under the same umbrella.

Recently, there has been a surge of research that both proposes and applies new approaches in numerous application areas, including Computer Vision (CV), Natural Language Processing (NLP), and Visual Question Answering (VQA). 
Despite the fact that EGL techniques are generally still in their early stage, the majority of existing studies have produced encouraging results, showing that the main DNNs can generally benefit from the additional explanation objective in terms of both model explainability and generalizability to unseen data across various application domains.
However, developing EGL frameworks can be difficult due to significant technical
% the invention and integration of related techniques to address the serious 
challenges caused by its unique characteristics, including:

% 1. EGL unique challenges (Challenges)
\begin{enumerate}
    \item \textbf{Gap between the pattern of model explanation and human explanation:} 
    The explanation generated by model explainers is typically continuous values, whereas human annotations are typically binary. Therefore, it is difficult to align the human explanation directly with the model explanation without significant efforts to fill the gap between the data domain and distributions.
    
    % \item \textbf{Difficulty in choosing the explanation technique:} The explanation-guided learning requires a differentiable model explanation in the first place, yet it is hard to decide what explanation method to extract the model explanation and apply the explanation guidance upon.
    % \textcolor{blue}{Moreover, it is still unclear how the explanation correctness correlates with the model generalizability on unseen data.}
    % whether the adjustment of the model explanation can lead to a faithful and generalizable explanation of unseen data.
    
    \item \textbf{Difficulty in comprehensively evaluating the EGL models:} unlike the conventional model where the task performance is the main focus, the quality of EGL outcomes generally needs sophisticated and carefully devised evaluation procedures that are often naturally subjective. For example, human participants can be involved in the evaluation to assess the quality of the model explanation.
    Moreover, beyond XAI explanation, EGL further requires the joint evaluation of the accuracy of prediction, explanation, and their mutual relation towards the reflection of model generalizability.
    Thus, we lack systematic standardization and comprehensive summarization approaches with which to evaluate the various EGL methodologies that have been proposed.
    
    \item \textbf{Noisiness in human annotation labels:}
    Unlike predictive task labels, it is much more likely for human annotators to unintentionally create noisy annotation labels where either the real important features are missed or irrelevant features are mistakenly included in the explanation annotation.
    For instance, when annotation the image data, some important object parts or even the entire objects may be missed by the coarsely drawn boundary from human annotators. Thus, applying naive supervision directly to train the model can lead to falsely excluded non-trivial features from the input space that are important to the prediction~\cite{gao2022res}.
    
    \item \textbf{Difficulty in explicitly measuring the faithfulness of the explanation quality with respect to the model generalizability:} Due to the fact that EGL techniques are generally still in their infancy, most existing works still primarily focus on merely evaluating the explanation quality of the EGL model independently of the model task performance. The faithfulness of the improved explanation quality with respect to the model prediction is yet to be explored explicitly and can be a key research question to be answered for EGL techniques to further advance and enhance the model performance and generalizability.
\end{enumerate}

\subsection{Contributions}
As the majority of existing EGL approaches were built for a specific application domain, cross-referencing these techniques across application domains serving different communities is problematic and challenging.
Moreover, the lack of a comprehensive review and taxonomy of existing techniques and applications in EGL creates substantial challenges for researchers working in the related field, since they lack clear information on existing bottlenecks, pitfalls, open-ended questions, and potentially fruitful future research directions.

To this end, this paper provides a systematic survey of EGL models across various application domains, including Computer Vision (CV)~\cite{41, 42, 43, 45, 50, 53, 55, 56}, Natural Language Processing (NLP)~\cite{7, 8, 9, 10, 11, 12, 13, 14, 15, 16, 17, 18, 19, 22, 23, 24, 26, 27, 29, 30, 31, 32, 33}, Visual Question Answering (VQA)~\cite{70, 71, 72, 73, 74, 75}, and more in Section 4. The goal of the survey is to help interdisciplinary researchers build a better understanding of the existing EGL techniques, and develop appropriate frameworks to solve the problems in their applications domains.
Besides, this survey also aims at helping researchers outside the AI communities to understand the basic principles as well as identify interdisciplinary open research opportunities in the EGL domain.
As far as we know, this is the first comprehensive survey on explanation-guided learning. This work's contributions are as follows:

\begin{itemize}
    \item We summarize a general learning paradigm of EGL based on existing works in this field to provide overall guidance on identifying and designing new EGL techniques.
    \item We identify the key factors in terms of comprehensively evaluating the EGL model's performance, and then provided a summarization and categorization of the existing evaluation procedures and metrics.
    \item We propose a taxonomy of explanation-guided learning categorized by the level of guidance and methodologies. The advantages, drawbacks, as well as relations among different subcategories of EGL techniques, are also introduced and compared.
    \item We introduce the broader application of EGL and detail the unique benefits and future opportunities for each application domain.
    \item We conduct a comprehensive experimental analysis and comparative study among existing EGL models in CV and NLP domains.
    \item We summarize the existing literature on EGL at the current stage, and then provide a set of open problems and potential promising future research directions of EGL.
\end{itemize}

\subsection{Relationship with Related Surveys}
This section outlines previously published surveys that have some relevance to Explanation-Guided Learning. These surveys can be classified into three topics: (1) XAI technique and evaluation, (2) AI ethics, and (3) interactive machine learning, as introduced in detail below.

\textbf{Explainablity Technique and Evaluation}: 
The related surveys of interpretability techniques provide a technical review and categorization of existing explanation techniques that can explain the machine learning model, especially for the sophisticated `black box' DNN models.
Several related surveys provide an in-depth classification of machine learning interpretability methods in general ~\cite{linardatos2020explainable, arrieta2020explainable, guidotti2018survey, roscher2020explainable}, while others focus on more specific fields of study. 
Specifically, Burkart et al.~\cite{burkart2021survey} review the explainability methods of supervised machine learning models.
Montavon et al.~\cite{montavon2018methods} provide a survey that specifically focuses on the interpretability techniques designed for explaining DNNs. Zhang et al.~\cite{zhang2018visual}  research interpretability techniques for Convolutional Neural Networks (CNN) and visual explanation. Tjoa et al.~\cite{tjoa2020survey}
summarized the XAI techniques that have been adopted for explaining medical data.
Along the line of interpretability techniques, many recent surveys also review the methods and metrics for comprehensively validating the quality of the explanation generated by the XAI techiniques~\cite{hoffman2018metrics, zhou2021evaluating, mohseni2021multidisciplinary}.

\textbf{AI Ethics}:
As the societal impact of AI grows, the goals for revising AI become more complex and diverse, ranging from improving a conventional model accuracy metric to infusing advanced human virtues such as fairness, accountability, transparency (FaccT), and unbiasedness~\cite{mao2019data}. 
Aligning to such direction, recent surveys started to collect, synthesize, and structuralize the existing approaches meant to be designed to handle several types of bias in AI~\cite{mehrabi2021survey, caton2020fairness}.
The most noteworthy finding in our survey for the landmark surveys is that the approaches for detecting bias in ML are more than the ways to mitigate the bias~\cite{bellamy2019ai, du2020fairness}.
The second important finding is that even though several studies focus on showing the ways to detect bias, they also present a hint of how we can mitigate them by showing some typical bias cases~\cite{mehrabi2021survey, ntoutsi2020bias}. Lastly, the existing survey also provides a pressing field needs explaining why we need to improve the ways to steer models in the case of witnessing the evidence of bias~\cite{hong2020human}.

\textbf{Interactive Machine Learning}:
Since Fails et al.~\cite{fails2003interactive} proposed the idea of interactive ML, the HCI community has put a high priority on applying XAI techniques in developing interactive techniques and systems meant to help ML engineers to better understand their models' weaknesses and strengths. Landmark surveys related to human factor and interaction can be categorized into 1) the interactive design--emphasizing how to design the feedback loop between humans and ML models through system~\cite{dudley2018review, jiang2019recent} that are widely proposed in the human factor research communities, such as SIGCHI, CSCW, and UIST, and 2) visual analytic--focusing on how to apply visualization techniques to help ML engineers understanding complex ML model behavior~\cite{yuan2021survey, endert2017state}.

\subsection{Outline of the survey}
The remaining part of the survey is organized as follows.
In Section 2, we introduce the problem formulation and performance evaluations of EGL models, as illustrated in Figure \ref{fig:gaps} and Table \ref{tab:eval}.
In Section 3, we provide the taxonomy of EGL categorized by the level of guidance and methodologies, as illustrated in Figure \ref{fig:tax}. 
Moreover, the details of each EGL technique, along with their corresponding advantages, drawbacks, and relations to other techniques in the same subcategories are provided.
In Sections 4 and 5, we first introduce the broader application of EGL and then conduct a comprehensive experimental analysis and comparative study among existing EGL models in both CV and NLP domains.
Lastly, we conclude the current development of EGL techniques and suggest several open problems and potential future research directions in Section 6.

\section{Problem formulation and performance evaluations}
This section begins by introducing the generic denotation and formulation of the Explanation-Guided Learning problem (Section \ref{sec:formulation}) and then considers ways to categorize the performance evaluation measures of Explanation-Guided Learning (Section \ref{sec:evaluation}).

\subsection{Problem formulation}
\label{sec:formulation}

Consider a differentiable model $f$ parameterized by $\theta$ that learns to fit inputs $X \in \mathbb{R}^{N\times D}$ and the corresponding one-hot class labels $Y \in \mathbb{R}^{N\times K}$, where $N$ denotes the total number of data samples, $D$ denotes the input dimension and $K$ denotes the number of classes.
An explainer $g$ is considered to extract the explanation $M$ from the model $f$ given its parameter $\theta$ and a set of data points $\langle X, Y \rangle$.
Generally speaking, the model explanation $M$ represents the marginal contribution of each input feature to the model's decision after all possible combinations have been considered.
Notice that in this paper we use the terms \textit{rationale}, \textit{attention}, and \textit{saliency maps} interchangeably as the specific form of $M$ that is frequently used by the corresponding application domains.
Depending on the way the explanation is calculated, $M$ can be generally represented by either local explanation $M^{(L)}$ where $M^{(L)}_i$ is the local explanation of model $f$ with respect to sample $\langle X_i, Y_i \rangle$, or a single global explanation $M^{(G)}$ of the model $f$.

\textbf{The Explanation-Guided Learning (EGL) paradigm.}
The general goal for Explanation-Guided Learning is to boost both the task performance as well as the interpretability of the backbone model by jointly optimizing model prediction as well as the explanation. 
Based on the earlier exploration of explanation supervision frameworks design \cite{4, 50, gao2021gnes, gao2022res}, we introduce the key objective function of Explanation-Guided Learning as follows: 

\begin{equation}
\label{eq:overall}
\min \ 
\underbrace{
\mathcal{L}_{\text{Pred}}(f(X),Y)
}_{\mbox{\small task supervision}}
+ 
\underbrace{
\alpha \mathcal{L}_{\text{Exp}} (g(f, \langle X, Y \rangle), \hat{M})
}_{\mbox{\small explanation supervision}}
+ 
\underbrace{
\beta \Omega(g(f, \langle X, Y \rangle))
}_{\mbox{\small explanation regularization}}
\end{equation}
where $\hat{M}$ explicitly incorporates the `right' explanation, which can be typically realized by human annotation masks \cite{7, gao2022aligning}. 

As shown in Equation \eqref{eq:overall}, the key objective function of Explanation-Guided Learning mainly consists of three terms, namely 1) task supervision term for the typical prediction loss (such as the cross-entropy loss), 2) explanation supervision term for supervising the model explanation with some explicit knowledge of what the `right' explanation should be, and 3) explanation regularization term for enforcing some general properties about the `right' explanation (such as maintaining the sparsity nature of the explanation).  Notice that all three terms above can be defined and implemented differently depending on each particular explanation-guided learning method.

\subsection{Performance evaluations}
\label{sec:evaluation}

\begin{figure*}
\centering
\includegraphics[width=0.95\linewidth]{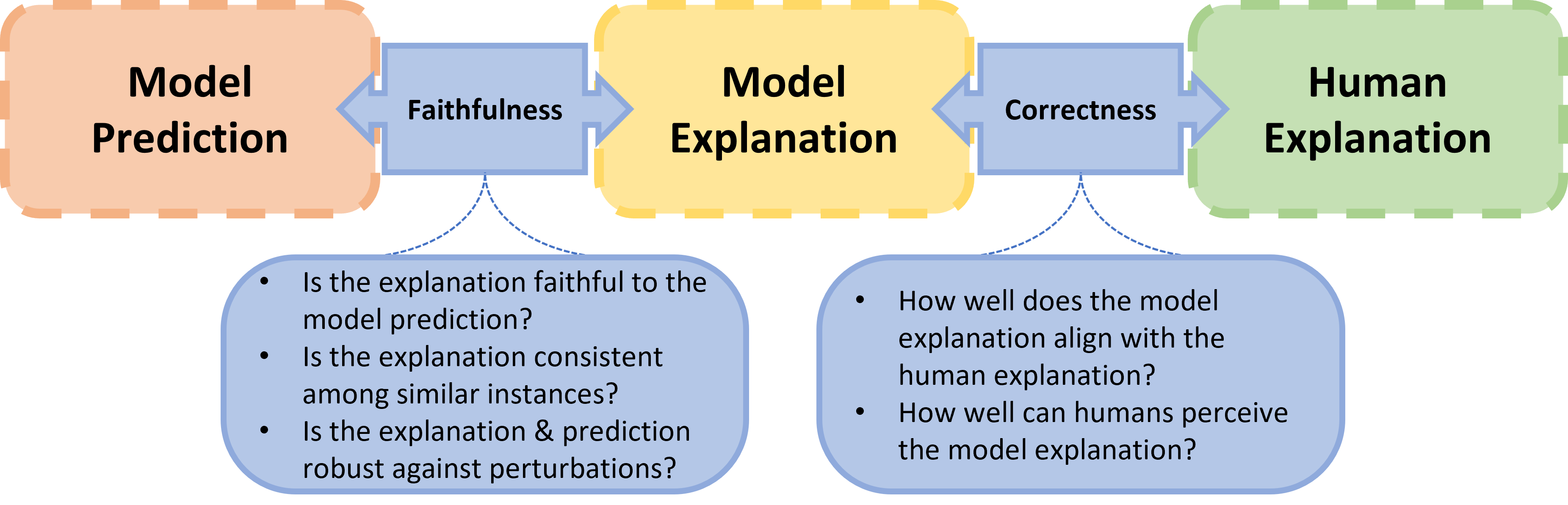}
\vspace{-5pt}
\caption{The gaps in Explanation-Guided Learning performance evaluations.}
\label{fig:gaps}
% \vspace{-0.3cm}
\end{figure*}

Unlike the evaluation of conventional machine learning models that typically only focus on the goodness of performance of the model, and the evaluation of traditional explainable AI models that only focus on the quality of the generated model explanation, Explanation-Guided Learning essentially jointly investigates the model prediction performance, the quality of model explanation, and their relation. 
As illustrated in Figure \ref{fig:gaps}, we identify and categorize two types of evaluations that are essential in measuring the performance of EGL models, namely the \textit{faithfulness} and \textit{correctness} of the model explanation.
Here we summarize the existing evaluation metrics into the two categories in Table \ref{tab:eval} and introduce each type of metric in great detail in the following two subsections.

\begin{table*}
  \caption{Detailed evaluation measures categorized by the gaps.}
  \centering
  \label{tab:eval}
\resizebox{0.9\textwidth}{!}{
  \begin{tabular}{c|c}
    \toprule
    Category & Evaluation Measure \\
    \hline
    \multirow{2}{*}{Explanation Faithfulness}
    &   Perturbation based \cite{7, 11, 25, 32, 50, 67, 68, 89}         \\
    &   Explanation Consistency based \cite{11, 89}         \\
    \hline
    \multirow{3}{*}{Explanation Correctness}
    &   Case study based \cite{13, 24, 25, 75} \\
    &   Human annotation based\footnote{These evaluation metrics will require human annotated ground truth explanation.} \cite{7, 9, 10, 16, 25, 31, 34, 35, 50, 56, 70, 72, 75, 89} \\
    &   User study based (user-perceived understandability) \cite{31, 32, 66, 86}  \\
    \bottomrule
  \end{tabular}
}
\end{table*}

\subsubsection{Metrics on evaluating explanation faithfulness} Here we introduce the metrics for explanation faithfulness (model prediction v.s. model explanation) evaluation, which aims at evaluating how the model-generated explanation influences the corresponding model's prediction.

\textbf{Perturbation-based evaluations:}
To evaluate the faithfulness of the model explanation, the study of how different types of perturbations on the input space influence the model prediction has become a very common and well-received approach in the literature \cite{7, 11, 25, 32, 50, 67, 68, 89}. Existing measures can be mainly categorized into three groups, depending on the type of perturbation as follows:

\begin{itemize}
    \item \textbf{Occlusion-based perturbation}: These metrics basically study how much influence on the model's prediction if the important feature or rationale identified by the model explanation are occluded or masked from the original sample \cite{7, 32, 50, 67, 68}. One commonly used occlusion-based metric is \textit{comprehensiveness} \cite{7}, where the difference of the predicted probability from the model $f(\cdot)$ for the same class $Y_i$ is compared between the original input $X_i$ and $X_i\backslash g(f, \langle X_i, Y_i\rangle)$, where the operation `$\backslash$' represents the exclusion of the supporting rationales $g(f, \langle X_i, Y_i\rangle)$ from input $X_i$. Mathematically, Comprehensiveness can be defined as follows:
\begin{equation}
\label{eq:comp}
\text{Comprehensiveness}  = f(X_i)_{Y_i} - f(X_i\backslash g(f, \langle X_i, Y_i\rangle))_{Y_i}
\end{equation}
    Besides the comprehensiveness score, many other intuitive methods are also used to evaluate the quality of the explanation. Inspired by previous works \cite{nguyen2018comparing, serrano2019attention}, a common intuitive strategy to measure the faithfulness of the explanation used by existing works \cite{32, 50, 67} is to track the degradation of model performance by removing importance features (often in decreasing order) from the input.

    % hold the detail description task for now.
    \item \textbf{Insertion-based perturbation}: These metrics study how well the prediction aligns between the original sample and an artificially generated sample where only the important feature/rationales are included \cite{7, 11, 50, 89}.
    One popular metric is Sufficiency \cite{7}, which captures the degree to which the snippets within the extracted rationales $g(f, \langle X_i, Y_i\rangle)$ are adequate for a model to make a prediction. Concretely, it can be defined as follows:
\begin{equation}
\label{eq:suff}
\text{Sufficiency}  = f(X_i)_{Y_i} - f(g(f, \langle X_i, Y_i\rangle))_{Y_i}
\end{equation}
    Similarly, many other intuitive methods are also used following the insertion idea. A common strategy used by existing works \cite{11, 50} is to track the increase in model performance by gradually inserting the important features (often in decreasing order) from the input.

    \item \textbf{Adversarial perturbation:} These metrics in general check whether the model explanation is still faithful to the model prediction under adversarial attacks \cite{25, 67}. For instance, \cite{67} leveraged the sanity check method originally proposed by \cite{adebayo2018sanity} to check if attribution maps look different when the deep network being explained is extremely perturbed or under adversarial attacks. The intuition behind this measure is that a faithful attribution method should yield different explanations for the randomized model. 
\end{itemize}

\textbf{Consistency-based evaluations:}
Besides the perturbation-based metrics which only focus on evaluating each instance locally at a time, existing works also propose consistency-based evaluation, where more global evaluation metrics have been proposed to validate how well the explanation aligns across similar instances \cite{11, 89}.
More specifically, \cite{11} proposed a metric called \textit{Data Consistency} that measures how similar the explanations for
similar instances are. Although the specific equation of the measurement in the paper is specifically designed for NLP and generative explanation, the basic idea can be generally expressed as follows:
\begin{equation}
\label{eq:DC}
\text{Data Consistency}  = |g(f, \langle X_i, Y_i\rangle) - g(f, \langle X_i \backslash M , Y_i\rangle)|
\end{equation}
where $M$ is a random mask that masks out $K$ input features from $X_i$; $K$ will be treated as a hyper-parameter depending on the dataset. In short, the general assumption behind this is that the model explanation between very similar samples should also be close to each other, so higher values represent better performance. Besides, the authors also suggested that it can also serve as an additional regularisation term during training for the model to be consistent in the generated explanations. 

Similar to the above idea, another work employed Intersection over Union
(IoU) score to measure explanation stability across similar instances \cite{89}. Specifically, they proposed to find similar instances by searching for the nearest neighbors of $X_i$ in the dataset based on both the semantic similarity – cosine of their BERT representations; and the lexical similarity – the ratio of overlapping n-grams. 

\subsubsection{Metrics on evaluating explanation correctness} Here we introduce the metrics for explanation correctness evaluation, which aims at evaluating how well the model-generated explanation aligns with the human explanation annotation or how well can humans perceive the model-generated explanation. 

\textbf{Case study:} Case study has been widely used as a conventional method for qualitatively evaluating the explanation generated by the model \cite{13, 24, 25, 75}, where a set of instances and their corresponding model explanations are selected and investigated qualitatively.
Although making qualitative assessments and detailed analyses of just a few samples can be easily achieved, it is in general less scientifically rigorous and the claims or conclusions are pruned to be biased due to the author's subjectivity.

\textbf{User study (user-perceived understandability):} User study, specifically user-perceived understandability, has been commonly used as a qualitative evaluation method to assess how humans can understand the explanation generated by the model \cite{22, 31, 66, 86}.
The user-perceived understandability methods are typically achieved by developing a user interface to show the model explanations to human subjects, and collecting the rating of how likely the important features identified by the model explanation can lead to the correct prediction of the underlying ground-truth label.

\textbf{Human annotation-based evaluation:} Explanation alignment is a unique yet commonly used quantitative metric in Explanation-Guided Learning which measures how the human-annotated ground truth explanation is aligned with the model generated explanation \cite{7, 9, 10, 16, 25, 31, 34, 35, 50, 56, 70, 72, 75, 89}. The distance is commonly measured by the Intersection over Union (IoU) score \cite{7, 16, 89}, precision, recall, and F-1 scores \cite{9, 89}.

\subsubsection{Other general metrics}
Besides measuring the faithfulness and correctness of model explanation, most of the papers also included the conventional model task performance metrics to verify if the Explanation-Guided Learning actually helped the generalizability of the backbone DNN models. Like most papers working on classification tasks, the common metrics used to evaluate model performance are accuracy, AUC (Area Under the ROC Curve) score, and F1 score.
\section{Explanation-guided learning techniques}

\begin{figure*}
\centering
\includegraphics[width=0.95\linewidth]{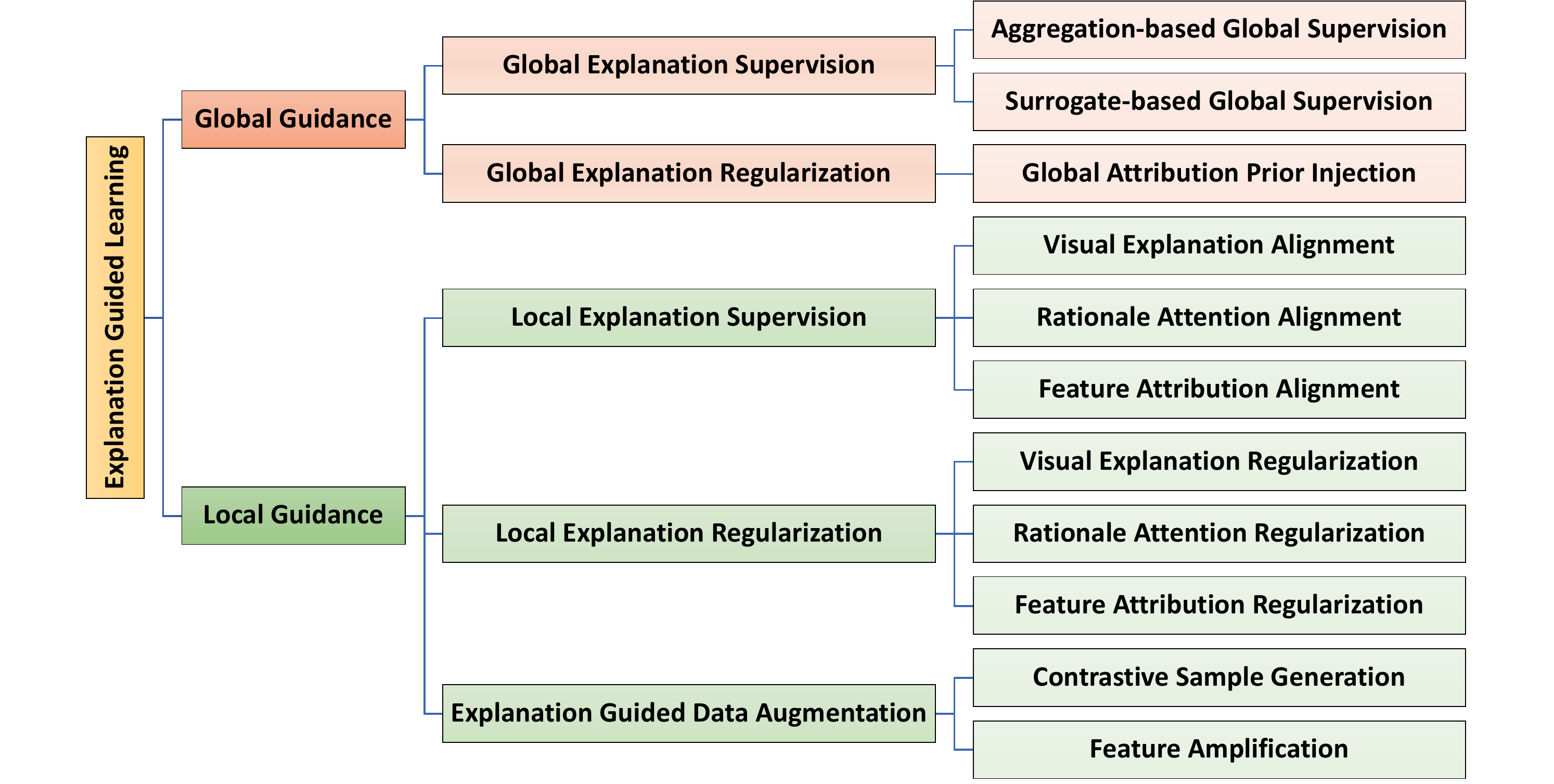}
\vspace{-5pt}
\caption{Taxonomy of Explanation-Guided Learning problems and techniques.}
\label{fig:tax}
% \vspace{-0.3cm}
\end{figure*}

This section focuses on the taxonomy and representative techniques utilized for each category and subcategory. According to the level at which the model explanation is obtained and supervised, the technique types for EGL can be divided into global guidance and local guidance, as shown in Figure \ref{fig:tax}. 
Specifically, global guidance focuses on the model's global explanation and refines the model's overall decision-making process, while local guidance guides the model with each sample-specific explanation.
The aforementioned techniques are then further categorized in terms of the way explanation guidance is injected during the course of model training.

\subsection{Global Guidance}
\begin{figure*}
\centering
\includegraphics[width=0.95\linewidth]{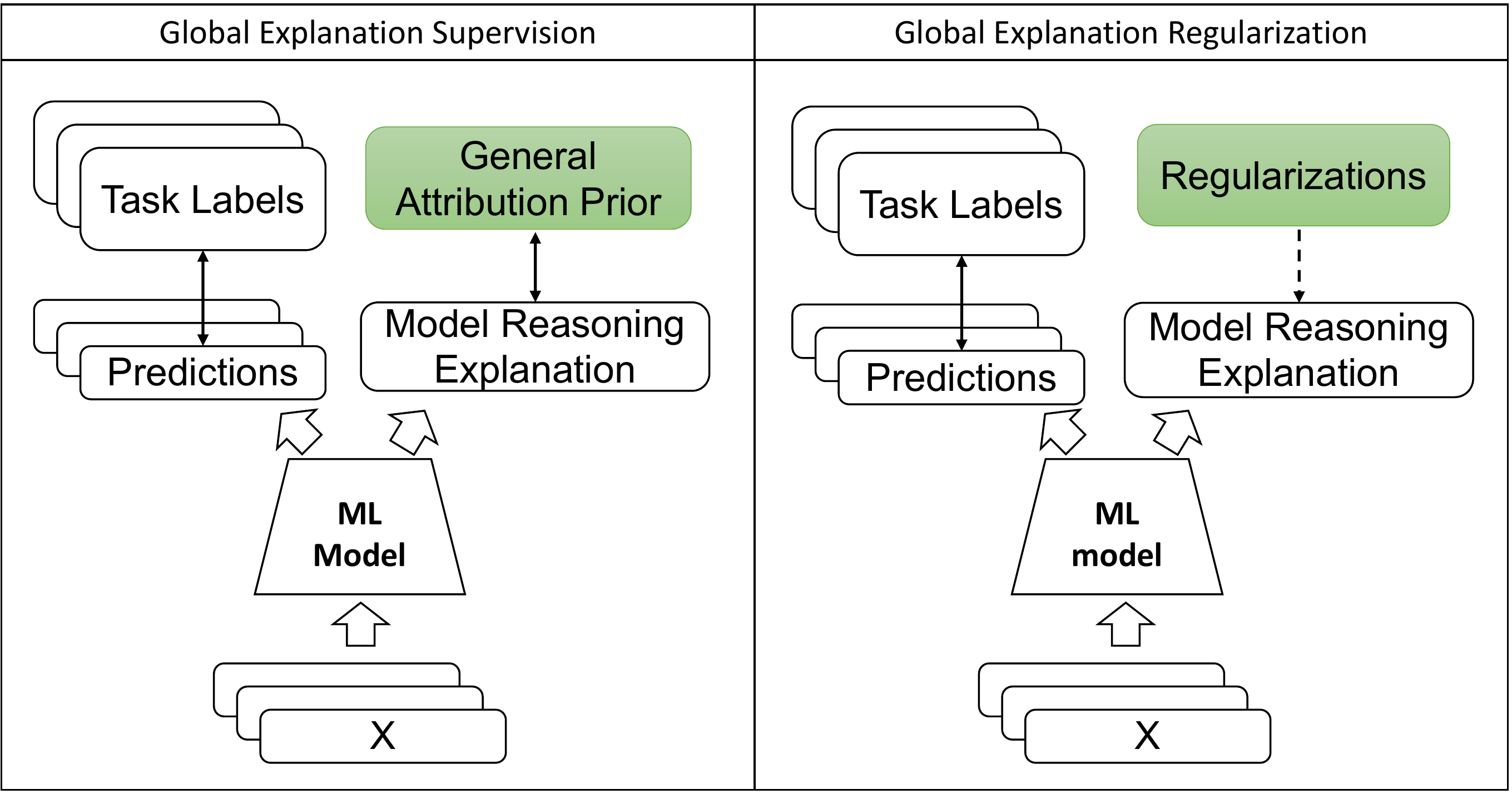}
\vspace{-5pt}
\caption{Illustration of Global Guidance techniques. Specifically, global explanation supervision techniques (left) aim at providing supervision in terms of model attribution, while global explanation regularization techniques (right) aim at confining the model reasoning process with prior knowledge. }
\label{fig:global}
% \vspace{-0.3cm}
\end{figure*}

Global explanation guidance focuses on injecting prior knowledge or adding supervision signals to improve the model's global explanation that explains the decision-making process of the model in general.
Based on the way explanation guidance is injected, global explanation guidance methods can be categorized into two types: 
1) Global Explanation Supervision: The ground truth explanation labels are provided as an additional supervision signal to train the feature-wise explanation of the model; and 2) Global Explanation Regularization: in which some regularization terms that represent some general prior knowledge about the model explanation are added to regularize the feature-wise explanation of the model, as illustrated in Figure \ref{fig:global}.

\subsubsection{Global Explanation Supervision}
The techniques proposed in global explanation supervision~\cite{15, 43, 53} aim at providing a single feature-wise explanation of the model globally.
Compared with instance-level local explanation supervision where the explanation ground truth is provided for each instance~\cite{50, gao2022aligning, gao2022res}, global explanation aims to provide a more effective global guide to the model's behavior as a whole.
% Different from a more commonly used instance-level local explanation supervision, where the explanation ground truth is provided for every instance~\cite{50, gao2022aligning, gao2022res}, the techniques proposed in global explanation supervision aim at providing a single explanation signal to the model feature-wise explanation to provide a more efficient global guide to the model's behavior overall~\cite{15, 43, 53}. 
Depending on the strategies to compute the global explanation of the model, current literature can be mainly categorized in two directions: 1) aggregation-based~\cite{15, 43} and 2) surrogate-based~\cite{vojivr2020editable, popordanoska2020machine, daly2021user}.

\textbf{Aggregation-based Global Supervision}:
This type of method typically achieves explanation supervision by first estimating the global feature attribution via aggregating local feature attribution of each sample and aligning it with a single ground truth feature attribution vector $\hat{m}$ as the additional supervision signal to train the model jointly with the conventional task loss. The common techniques used to calculate each sample's feature attribution are integrated gradient~\cite{15} and the expected gradient proposed by Erion et al.~\cite{43}.
Specifically, the objective function for aggregation-based global supervision can be summarized as follows:
\begin{equation}
\label{eq:global_ES}
\min \ 
\mathcal{L}_{\text{Pred}}(f(X),Y)
+ 
\alpha \mathcal{L}_{\text{Exp}} (\frac{1}{N}\sum^N_{i=1} g(f, \langle X_i, Y_i \rangle), \hat{M})
\end{equation}

This type of method has been utilized and shown promising results in many application domains, such as text classification~\cite{15} and image classification tasks~\cite{43}.
The advantage of the aggregation-based global supervision methods is that they can easily adapt existing techniques developed for local explanation with little to no extra effort. 
Besides, the acquisition of only one single feature attribution vector as a class-wise explanation signal is much more affordable, as compared with instance-wise supervision methods which require much more labor from human annotators.
However, the drawbacks of this type of technique also come from the aggregation of local explanation, as the aggregated explanation is sensitive to the samples used to calculate, and thus could bring the sample bias into the global explanation of the model estimated.

\textbf{Surrogate-based Global Supervision}:
This branch of work achieves explanation supervision by first estimating the global explanation of the target model via a surrogate model where the model-level explanation is easy to obtain, and then human knowledge can be leveraged to guide the global explanation and consequently supervise the model behavior.
In this branch, the rule-based explanation is commonly used as it can be easily understood and edited by practitioners~\cite{vojivr2020editable, popordanoska2020machine, daly2021user}. 

Rule-based explanation supervision can be achieved from many different angles.
For instance, Voj{\'\i}{\v{r}} et al.~\cite{vojivr2020editable} proposed the editable rule-based models that enable the users to edit rules and replace the underlying machine learning model; 
Popordanoska et al.~\cite{popordanoska2020machine} proposed the Explanatory Guided Learning (XGL) framework that creates simple rules capturing the prediction of the target model and allows the user to correct instances that are incorrect and the model is retrained.
Besides, the rule-based explanation is also used as a mechanism for feedback that supports user adjustments without retraining the model~\cite{daly2021user}; Cornec et al.~\cite{cornec2021aimee} developed the AI Model Explorer and Editor tool (AIMEE) that provides visualization of model decision boundaries using interpretable surrogates, and allows for the real-time modification of the decision boundaries.
More recently, Lee et al.~\cite{lee2022self} proposed SELOR, a framework for upgrading a deep model with a Self-Explainable version with LOgic rule Reasoning capability, inspired by neuro-symbolic reasoning~\cite{de2020statistical} that integrates deep learning with logic rule reasoning to inherit advantages from both. SELOR provides high human precision by explaining logic rules while also maintaining high prediction performance, and does not require predefined rule sets and can be learned
in a differentiable way.

\subsubsection{Global Explanation Regularization}
Global explanation regularization is the method where some regularization terms that incorporate general prior knowledge about the global explanation are applied to the model.
A good example of a preferred property of the model explanation is the sparseness, as it can provide a better understanding of the model behavior by humans, and in the meantime, serve as a regularizer of the explanation space to enhance model generalizability~\cite{setiono1997neural, peng2009lazy, SMART}.
Concretely, the objective function for global explanation regularization can be summarized as follows:
\begin{equation}
\label{eq:global_ER}
\min \ 
\mathcal{L}_{\text{Pred}}(f(X),Y)
+ 
\beta \Omega(M^{(G)})
\end{equation}
where function $\Omega(\cdot)$ represents the specific regularization function for regulating the model's global explanation, and $M^{(G)}$ represents the model's global explanation vector calculated based on intrinsic parameters of the model $f$.

A commonly used prior knowledge to define $\Omega(\cdot)$ is to ensure the sparseness of the explanation, where a regularization term is proposed to penalize small magnitude weights of $f$ that connect to the input features~\cite{setiono1997neural, peng2009lazy, SMART}.
The existing studies suggest that this can result in a feature selection effect, and greatly enhance the model's computational efficiency as well as generalizability.
In addition, Burkart et al.~\cite{burkart2020batch} proposed a batch-wise regularization technique to enhance the interpretability of DNN models by means of a global surrogate rule list with a novel regularization approach that yields a differentiable penalty term.
In Wu et al.~\cite{wu2020regional}, the authors proposed the regional tree regularization that encourages a DNN model to be well-approximated by several separate decision trees specific to predefined regions of the input space, yielding simpler explanations without compromising model accuracy.

\subsection{Local Guidance}

\begin{figure*}
\centering
\includegraphics[width=0.95\linewidth]{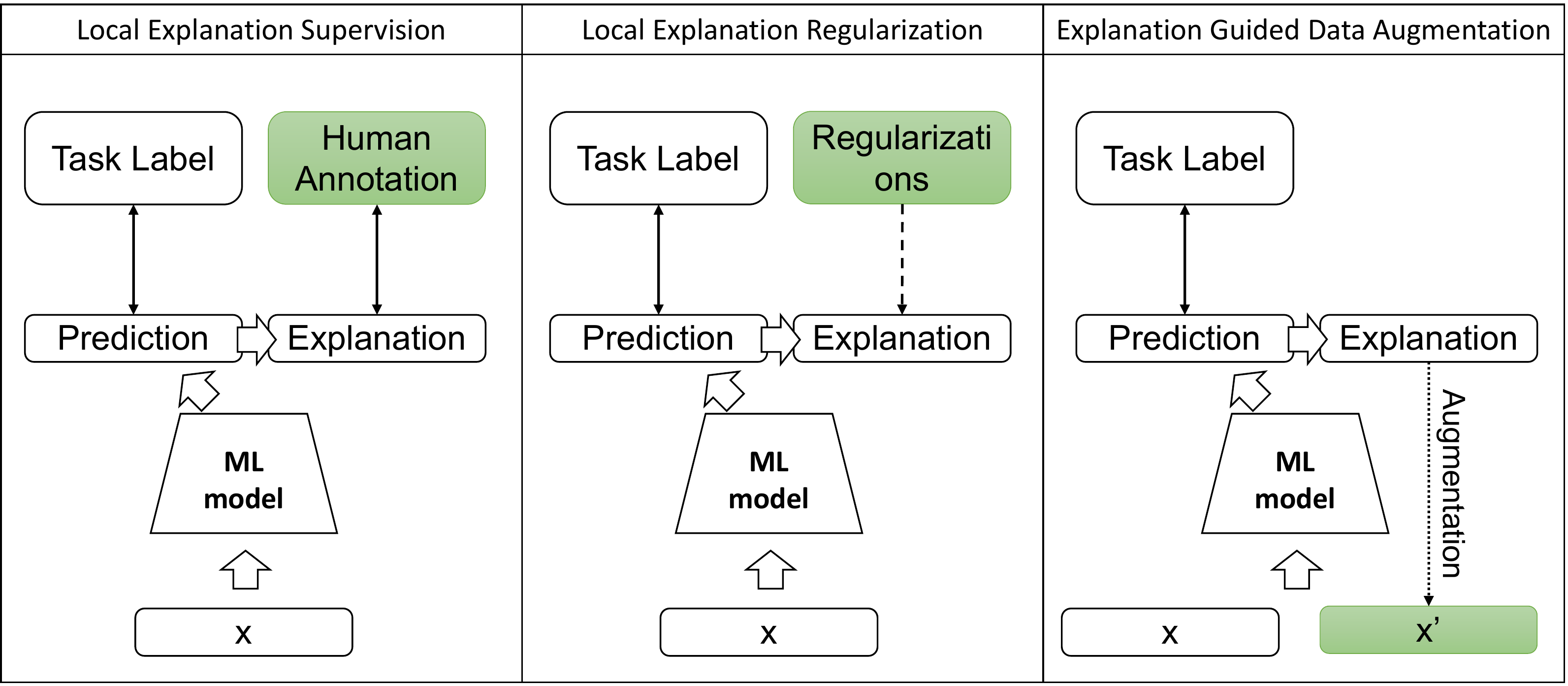}
\vspace{-5pt}
\caption{Illustration of Local Guidance techniques. (left) Local Explanation Supervision: The ground truth explanation labels for each individual sample are provided as additional supervision signals to train the corresponding model explanation; (middle) Local Explanation Regularization: in which some regularization terms that represent some general prior knowledge about the local model explanation are added to regularize all the local explanation of the model; (right) Explanation Guided Data Augmentation: where the local model explanations are used to construct additional data samples for model training.}
\label{fig:local}
% \vspace{-0.3cm}
\end{figure*}

Local explanation guidance focuses on applying supervision signals or regularization terms to the model explanation of each local sample to guide the model learning. 
As shown in Figure \ref{fig:local}, compared with the global explanation guidance, local guidance is more commonly used and explored in the current research communities thanks to the development of local explanation techniques, such as GradCAM \cite{selvaraju2017grad} and attention mechanism \cite{bahdanau2014neural, vaswani2017attention}.
Based on the way explanation guidance is injected, local explanation guidance techniques can be categorized into three types: 
1) Local Explanation Supervision: The ground truth explanation labels for each individual sample are provided as additional supervision signals to train the corresponding model explanation; 2) Local Explanation Regularization: in which some regularization terms that represent some general prior knowledge about the local model explanation are added to regularize all the local explanation of the model; and 3) Explanation Guided Data Augmentation: where the local model explanations are used to construct additional data samples for model training.

\subsubsection{Local Explanation Supervision}
Just as we supervise the model prediction via ground truth labels, local explanation supervision methods add additional supervision signals to align the model explanation with ground truth explanation labels (e.g. human annotation masks) during model training. The explanation loss and the conventional prediction loss are typically jointly optimized during model training.
The general assumption behind this approach is that the model can benefit from the explanation labels by learning to focus on the right features and consequently lead to better generalizability to unseen instances.
Depending on the data representation and application domains, we further narrow down the techniques into three subcategories:
1) visual explanation alignment, 2) rationale attention alignment, and 3) feature attribution alignment.

\textbf{Visual Explanation Alignment:} 
The visual explanation of image data is typically represented by a heat map overlaid on top of the original image, and the ground truth explanation labels $\hat{M}$ are typically obtained by human annotation in the form of bounding boxes or fine-grained contours.

The first framework that can be applied to visual explanation alignment was proposed by Ross et al.~\cite{4},  
where the authors defined a very generic explanation-guided learning loss called ``Right for the Right Reasons'' loss (RRR) as follows:
\begin{equation}
\label{eq:rrr}
\min \ 
\sum_{i=1}^N -Y_i \log (f(X_i))
+ 
\alpha \sum_{n=1}^N (\hat{M}_{i} \frac{\partial}{\partial X_{i}} \log (f(X_{i})))^2
+ 
\beta \|\theta\|_2^2
\end{equation}
where $\hat{M}_{i}$ denotes the ground truth explanation mask of a sample $i$; the task supervision loss is implemented as the conventional cross-entropy loss, and the explanation supervision loss is designed to enforce the alignment of the ground truth explanation mask $\hat{M}$ and the gradient maps via inner product operations.

Later on, the RRR loss is further extended by Schramowski et al.~\cite{5} and Dharma et al.~\cite{45} regarding the definition of the explanation losses. 
Specifically, instead of regularizing the gradients with respect to input $X$, Schramowski et al.~\cite{5} proposed to regularize the gradients of the final convolutional layer of the model that corresponds to GradCAM explanation and add a rescaling weight $c_k$ to each class $k$ for handling the unbalanced dataset issue. 
% Concretely the explanation supervision loss in Schramowski et al.~\cite{5} is shown below:
% \begin{equation}
% \label{eq:rrr-ext1}
% \mathcal{L}_{\text{Exp}}  = 
% \sum_{i=1}^N (\hat{M}_{i} \frac{\partial}{\partial h_{i}} \sum_{k=1}^K c_k \log (f(X_{i})))^2
% \end{equation}
In Dharma et al.~\cite{45}, the explanation loss is broken down into two terms to characterize the sensitivity of the gradient maps differently based on the relationship between each pixel of input and the ground truth mask, as shown below:
\begin{equation}
\label{eq:rrr-ext2}
\mathcal{L}_{\text{Exp}}  = 
\alpha_1 \sum_{j \in \hat{M}_i} \frac{\partial
\mathcal{L}_{\text{Pred}}(f(X_i),Y_i)}{\partial X_{i,j}} 
+
\alpha_2 \sum_{j \in [d] \backslash \hat{M}_i} \frac{\partial
\mathcal{L}_{\text{Pred}}(f(X_i),Y_i)}{\partial X_{i,j}} 
\end{equation}
where $[d] \backslash \hat{M}_i$ represent the complement subset of the explanation $\hat{M}_i$ of the whole feature set.

Later, many more models that are designed for visual explanation alignment are proposed \cite{41, 50, ying2022visfis, gao2022aligning, gao2022res}. Specifically, Stammer et al.~\cite{41} proposed a visual explanation alignment model based on symbolic (concept) alignment, where The symbolic (concept) explanation is modeled by a set transformer module.
Mitsuhara et al.~\cite{50} proposed a visual explanation alignment objective specifically designed for the Attention Branch Network (ABN) \cite{fukui2019attention}, where the attention branch outputs are used as the model explanation. The limitation of this work is that it can only work under ABN architecture.
Ying et al.~\cite{ying2022visfis} proposed the Visual Feature Importance Supervision (VISFIS) framework that optimizes four key model objectives: (1) accurate predictions given limited but sufficient information (Sufficiency); (2) max-entropy predictions given no important information (Uncertainty); (3) invariance of predictions to changes in unimportant features (Invariance); and (4) alignment between model explanations and human explanations (Plausibility) to improve model
accuracy as well as performance. 
Nguyen et al.~\cite{nguyen2022visual} proposed two novel architectures of self-interpretable image classifiers that first explain, and then predict by harnessing the visual correspondences between a query image and exemplars, and demonstrated the improvement on out-of-distribution (OOD) dataset scenarios.
Gao et al.~\cite{gao2022aligning} proposed a more generic visual explanation alignment framework called GRADIA based on the GradCAM explanation. In addition, they proposed the Reasonability Matrix that can better determine what samples need to be adjusted to improve the model performance and explanation quality. 
More recently, Gao et al.~\cite{gao2022res} proposed a robust visual explanation alignment framework that can better handle the nosiness of human annotation on image data. 
Specifically, the explanation loss is defined as follows:
\begin{equation}
\label{eq:res}
\mathcal{L}_{\text{Exp}}  = 
    \min_{\theta, \lambda, \phi}
    \sum\nolimits^N_i
    \max\{0,\|\tilde{M}_i - \hat{M}_i\|-\lambda\}
    +(g(f, \langle X_i, Y_i \rangle) - h_\phi(\hat{M}_i))^2
\end{equation}
where $\phi$ is the parameter set of the imputation function $h_\phi(\cdot)$. The imputation function can be realized by applying multiple layers of convolution operations with learnable kernels over the raw annotation labels; $\tilde{M}_i$ is a binary projection of the explanation $g(f, \langle X_i, Y_i \rangle)$ by a threshold $\lambda$, as:
\begin{equation}
\label{eq:M_hat}
\tilde{M}_i = \left\{
        \begin{array}{ll}
            1  & \quad g(f, \langle X_i, Y_i \rangle)\geq \lambda \\
            -1 & \quad g(f, \langle X_i, Y_i \rangle)<\lambda \\
        \end{array}
    \right.
\end{equation}

Besides the application to general images, visual explanation alignment techniques have also been applied to medical image domains \cite{47, 57, 58}. Please refer to Applications Section for more details.

\textbf{Rationale Attention Alignment:} The explanation of natural language data is typically represented by the rationales (e.g. word tokens) that highlight the most significant part of the data for making specific task predictions. The ground truth explanation labels are typically obtained by human annotation in the form of rationales or natural language format (sentences). 
In this domain, the datasets collected by the ERASER benchmark \cite{7} are commonly used as the datasets come with ground truth rationales obtained from human annotators.

Many existing works have proposed to supervise the rationale attention of the model to improve the model performance and quality of attention \cite{9, 10, 13, 18, 21, 24, 33}. 
The explanation loss is commonly realized by conventional losses, such as cross-entropy loss \cite{21, 24}, Mean Squared Error (MSE) \cite{26, 33}, and KL-divergence loss \cite{18}. Besides, Atanasova et al.~\cite{11} proposed several novel ways to enforce the alignment, such as Data Consistency, Confidence Indication, and Faithfulness. 

In addition to leveraging the model attention value itself, many existing works have also proposed to directly generate rationales \cite{12, 18, 26, 27, 30, 31} or natural language \cite{16, 17} as the `explanation' of the model to be aligned with ground truth labels via additional decoders, such as Conditional Random Field (CRF)~\cite{lafferty2001conditional, 12}, Gated Recurrent Unit (GRU)~\cite{chung2014empirical, 18}, Transformer-based models~\cite{vaswani2017attention, devlin2018bert, 26, 16}.

\textbf{Feature Attribution Alignment:} Besides the specific domain of applications, local explanation supervision can be generally applied to any dataset where the input feature importance can be computed. Such feature importance is typically referred to as `Feature Attribution', and can be also treated as the model explanation to be aligned with human explanation ground Truth. 
For instance, in Balayan et al.~\cite{80} the feature attribution is computed by a specific designed semantic layer (an intermediate output that is the importance of each feature) and is aligned with human-labeled feature masks by Cross-Entropy loss;
in Singh et al.~\cite{81}, feature attribution is calculated by Contextual Decomposition, and is aligned with the ground truth human labels by $\ell_1$ distance~\cite{42}.

Overall, the idea of local explanation supervision has been explored extensively in many application domains in recent few years, primarily due to the fact that 1) it is straightforward for human annotators to provide an instance-wise explanation with necessary domain knowledge, and 2) the development and popularity of local explanation techniques, such as GradCAM \cite{selvaraju2017grad} and attention mechanism \cite{bahdanau2014neural, vaswani2017attention} to explain complex DNNs in high dimensional problem space (such as image and text data).
So far the results seem to be promising, as most existing works suggest that applying the local explanation supervision during training can greatly enhance both the quality of the explanation as well as the performance of the backbone DNNs model.
However, as also pointed out by several existing works, the scalability remains the biggest challenge for this kind of approach, as the additional instance-wise human explanation labels may not be easily accessible and require non-trial effort from human annotator~\cite{gao2021gnes, gao2022res}.
Designing effective semi-supervised or weakly-supervised explanation supervision frameworks, or even adapting the idea from active learning can be promising future directions to further overcome this limitation.
Nevertheless, existing works also demonstrated the effectiveness of local explanation supervision under very limited training sample sizes~\cite{gao2022res, gao2022aligning}, which could suggest the potential benefit of applying current techniques to the domains where data samples are limited and hard to acquire, yet both model performance and the explainability are on-demand, such as in medical domains.

\subsubsection{Local Explanation Regularization}

%%TODO try equation
Local explanation regularization methods add additional regularization terms to regularize each local explanation to ensure the generated model explanations follow some general properties (such as smoothness, stability, and sparsity) or follow the knowledge from other existing well-trained models. 
The additional explanation of regularization loss is typically jointly optimized with the prediction loss during model training.
Depending on the type of regularization terms, we break down the existing techniques into two subcategories:
1)Property-based Regularization and 2) Explanation Distillation Regularization.

\textbf{Property-based Regularization:} The additional regularization terms are injected into the model explanation to enforce some general properties (such as smoothness, stability, and sparsity). Specifically, in Lei et al.~\cite{27} the authors proposed the \textit{continuity} and \textit{sparsity} regularization terms;
In Erion et al.~\cite{43} the authors proposed the \textit{smoothness} Regularization (i.e. Laplace 0-mean prior) on the model Explanation computed by expected gradient;
In Halliwell et al.~\cite{64} the authors proposed the Prediction-guided sparsity regularization (in Equations (5) and (6)) to penalize the model to have small values in saliency maps (computed by GradCAM and guided BP) if the prediction is incorrect;
Alvarez et al.~\cite{62} proposed a gradient regularization approach for enforcing explanation \textit{robustness/stability};
Plumb et al.~\cite{82} apply the \textit{fidelity} and \textit{stability} regularization on the explanation.	Specifically, the explainer $g(\cdot)$ is realized by Local Interpretable Model-Agnostic Explanations (LIME)~\cite{ribeiro2016should} with a linear function $l(\cdot)$, and the authors applied two regularization terms on the model explanation: 1) neighborhood-fidelity and 2) stability  based on the neighborhood of input, as shown below:
\begin{equation}
    \Omega= 
    \underbrace{
    \mathbb{E}_{X^\prime \sim \mathcal{N}_{X_i}} [(l(X^\prime) - f(X^\prime))^2] 
    }_{\mbox{\small neighborhood-fidelity}}
    +
    \underbrace{
    \mathbb{E}_{X^\prime \sim \mathcal{N}_{X_i}} [\| g(f, \langle X_i, Y_i \rangle) - g(f, \langle X^\prime, Y_i \rangle) \|_2^2]
    }_{\mbox{\small stability}}
\end{equation}
where $\mathcal{N}_{X_i}$ is a neighborhood of sample $X_i$ in the space of probability distributions over the whole input data distribution $X$, and $X^\prime$ is sampled from the neighborhood $\mathcal{N}_{X_i}$.
Intuitively, the fidelity regularization enhanced the explanation to accurately convey which patterns the model used to make this prediction, while the stability regularization will lead to more stable explanations, which will improve the model's trustworthiness~\cite{62, alvarez2018robustness}.

\textbf{Explanation Distillation Regularization}: Besides enforcing predefined properties of the explanation, this line of work tries to distill explanation knowledge from other well-trained models to guide the explanation of the target model. 
In Zeng et al.~\cite{67}, the authors proposed to align the explanation of a target model with another pre-trained adversarially counterpart model generated explanation using $\ell_2$ distance loss.
Singh et al.~\cite{63} proposed to align the target model's Class Activation Maps (CAM)~\cite{zhou2016learning} explanation with a pre-trained model's explanation by minimizing the overlap between each classes explanation.
More specifically, the explanation loss consists of two terms, 1) regularization loss which measures the distance between the target models' and the corresponding pre-trained model's explanation of class $i$, and 2) overlapping loss which calculates the similarity between the target model's explanation of different classes.
As a result, the model can be trained under the constraints that 1) the target model explanation should be as close as possible to a pre-trained model, and 2) the model explanation of different classes should be as different as possible, leading to a batter explanation quality and higher accuracy.
More recently, Fernandes et al.~\cite{fernandes2022learning} proposed Scaffold-Maximizing Training (SMaT) framework for directly optimizing explanations of the model’s predictions to improve the training of a student simulating the said model. The authors found that, across tasks and domains, explanations learned with SMaT both lead to students that simulate the original model more accurately and are more aligned with how people explain similar decisions.

While using the idea from model distillation to extract the knowledge to guide the model explanation is an interesting direction, the potential positive effect is largely dependent on the choice and quality of the pre-trained model and is prone to negative transfer, such as contextual bias in pre-trained model explanation, that can hurt the target model performance. Thus additional validation and guidelines are on demand for this type of technique to be applied to handle real-world problems.

\subsubsection{Explanation Guided Data Augmentation.} Explanation Guided Data Augmentation is an emerging subdomain in the data augmentation domain, where the ground truth explanation (i.e. rationale) of the prediction task is taken into account when building up additional augmented samples for model training. 
The general formulation for generating explanation-guided data augmentation samples can be summarized as follows:
\begin{equation}
    X'_i = \text{aug} (X_i , g(f, \langle X_i, Y_i\rangle))
\end{equation}
Where $\text{aug}(\cdot)$ denotes the specific augmentation function based on the original input sample $X_i$ and the model's explanation for the given input-output pair $\langle X_i, Y_i\rangle$.

The underlying assumption is that training the model with the augmented samples $X'$ will encourage the model to better learn to pay attention to the right rationales for the prediction tasks and thus naturally enhances both the explainability as well as the generalizability of the model.
Based on the way explanation is used for the data augmentation, existing techniques can be categorized into two directions: 1) Rationale Inclusion/Amplification and 2) Rationale Exclusion/Masking.

\textbf{Rationale Inclusion/Amplification}: This line of works typically emphasizes the right rationales and de-emphasize other irrelevant features. The inclusion/amplification-based augmentation function can be generally defined as follows:
\begin{equation}
\label{eq:amp}
    \text{aug}_{in} (X_i , g(f, \langle X_i, Y_i\rangle)) = X_i \times (\gamma + \lambda g(f, \langle X_i, Y_i\rangle)))
\end{equation}
where $\gamma$ is used to set a default offset value to preserve all the feature values regardless of the importance; $\lambda$ is the scale factor that controls the degree of amplification of the important features.

Specifically, Sharma et al.~\cite{20} proposed to amplify the feature values of the right rationales relatively higher by a certain degree. 
In their experiments, $\gamma$ is set to 0.01, and $\lambda$ is set to 1 to emphasize the rationale features in the augmented samples.
The results demonstrated the general effectiveness of the proposed method on several conventional ML models, such as Naive Bayes, logistic regression, and SVM.
In Saha et al.~\cite{59} only the important part of the image for network prediction is selected using saliency-based explanations and stored in the episodic memory with the corner coordinate for continual learning.
Ismail et al.~\cite{68} proposed to minimize the KL divergence between $f(X)$ and $f(X')$, where $X'$ is augmented by masking the features with low gradient values.
These types of methods can be seen as a special case of Equation \eqref{eq:amp} where $\gamma$ is set to 0, and $\lambda$ is set to 1 to only include the rationale features in the augmented samples.

Besides the simple augmentation of the feature values, several other works have also proposed some novel and unique ways to augment data to best leverage the extra information from the model explanation.
In Pillai et al.~\cite{66}, the saliency map explanation of the original sample and other samples as a composed image is aligned with the model explanation of the original input sample.
In Teso et al.~\cite{86}, the irrelevant features are perturbed while the rationales and task labels are preserved as new samples to guide the model attending the ground truth rationales.
In Schneider et al.~\cite{61}, the model explanation (generated by GradCAM) is treated as additional input for model prediction and requires a specific change of the model architecture.
In summary, the general idea stays the same, which is to build additional samples and inform the model to better learn which features are the right rationales to make the right prediction of the downstream tasks.

\textbf{Rationale Exclusion/Masking}: As opposite to inclusion/amplification, this line of works typically teaches the model not to attend irrelevant rationales by excluding/masking out the right rationales, as summarized by the following equation:
\begin{equation}
\label{eq:mask}
    \text{aug}_{ex} (X_i , g(f, \langle X_i, Y_i\rangle)) = X_i \times (\gamma - g(f, \langle X_i, Y_i\rangle)))
\end{equation}
where $\gamma$ is typically set to be the maximum possible value of the importance, e.g. 1, to exclude the value of the important features from $X$, and thus serve as a masking function for the data augmentation.

Specifically, in Zaidan et al.~\cite{8}, the authors propose to construct some additional samples by masking out those important features of some existing samples to simulate the loss of confidence (uncertainty should raise) in predicting the right answer. 
A similar idea can be also found in Li et al.~\cite{56}, where the proposed self-guidance is basically using the model explanation as a mask to augment the original image and thus construct an unsupervised loss based on the augmented/masked image.

Overall, the unique advantages of explanation-guided data augmentation techniques can be summarized as follows:
1) it takes the model behavior (i.e. rationale for the prediction) into consideration;
2) it can be model agnostic with respect to the specific explainability techniques used for calculating the model explanation;
3) it can be used in combination with other conventional data augmentation techniques, and in parallel with other EGL techniques for model training.
However, the effectiveness of the existing works is mainly supported by intuitions and empirical observations.
Thus further development of quantitative evaluation metrics as well as theoretical analysis and justification of the techniques can be essential to further advance this field of research.

\section{Applications}

\subsection{Computer Vision (CV)}
Applying EGL to solve image classification problems has become a hot and attractive research area in recent years~\cite{50, gao2022aligning, gao2022res}, largely thanks to the popularity and advancement of visual explanation techniques~\cite{zhang2018visual, selvaraju2017grad}.
Depending on the nature of the image source, existing works can be further categorized into (1) general image prediction and (2) medical image analysis.

\subsubsection{General Image Prediction}
The application of EGL on general images typically involves image classification tasks on natural image data such as ImageNet~\cite{krizhevsky2017imagenet}, Caltech-UCSD Birds (CUB)~\cite{wah2011caltech}, Microsoft COCO~\cite{lin2014microsoft}, and Places365~\cite{zhou2017places}, and some synthetic image data such as ToyColor~\cite{4}, MNIST~\cite{lecun1998gradient}, and many MNIST variants including Fashion-MNIST~\cite{xiao2017fashion}, Decoy-MNIST~\cite{4}, and Color-MNIST~\cite{li2019repair}.
The typical EGL technique used in this application domain is local explanation supervision and regularization, where the sample level visual explanation of the model is jointly optimized together with the conventional prediction loss~\cite{gao2022aligning, gao2022res, 41, 42, 43, 45, 50, 53, 55, 56}.
When applying the explanation supervision techniques, the ground truth explanation labels are typically collected from human annotators, and the additional attention loss is typically realized by a distance loss between the ground truth and the model visual explanation at the sample level. 
For model explanation assessment, case studies are most commonly used for qualitative analysis~\cite{50, gao2022aligning, gao2022res}, while IoU score is for quantitative evaluation~\cite{56, gao2022aligning, gao2022res}.

\subsubsection{Medical Image Analysis}
Besides generic image applications, EGL has also been widely studied in the medical domain, thanks to the availability of domain-expert annotation on many medical image datasets~\cite{codella2018skin, langerhuizen2020deep, wang2017chestx}.
In general, we observed a variety of datasets studied by existing works, including but not limited to ISIC Skin Cancer dataset~\cite{codella2018skin}, Iris-Cancer dataset~\cite{lichman2013uci}, scaphoid fracture detection dataset~\cite{langerhuizen2020deep}, Fundus image dataset (IDRiD)~\cite{h25w98-18}, and the pneumonia detection X-ray dataset~\cite{wang2017chestx} for disease identification task~\cite{58}.
Similar to most EGL frameworks on generic image data, an additional explanation loss is added to the model objective and is typically realized by a distance loss between the ground truth annotation collected from domain experts and the model visual explanation. 
However, compared with generic image data, several unique challenges have been identified by existing works when applying EGL to medical images, such as 1) difficulty in assessing the quality of the model explanation, and 2) the scalability of the sample size of the annotation labels of the datasets.

\subsection{Natural Language Processing (NLP)}

% Explanation-Guided Learning
Interest has recently grown in applying EGL to designing NLP systems. Based on how the explanation is acquired, we have two categories of the application: (1) using the attention mechanism as the explanation and (2) using a generative model to generate the explanation.

\subsubsection{Attention mechanism as the explanation}
% EGL using attention mechanism as explanation}: refer to \cite{7, 8, 9, 10, 11, 13, 14, 15, 18, 24, 29, 32, 33}
%rationale-level annotation
% token-level
% To identify textual spans that determine the target label, which is known as faithful rationales, 
NLP systems generally use variants of attention mechanisms to get explanations. To evaluate the explanation, the ground truth explanation labels are typically collected from human annotators (Stacey et al. \cite{33} use TextRank to get ground truth labels), and the evaluation metric can be the F1 score and IoU score based on token or snippet level. In addition to the agreement with human rationales, a faithful explanation is related to the downstream task performance, so rationale-level supervision is widely applied \cite{13,14,18,24,29,33}. Comprehensiveness and sufficiency are two main metrics regarding the influence of the explanation on the downstream task, Faithfulness, Data Consistency, and Confidence Indication are other diagnostic properties \cite{11}.
Attention mechanisms can learn to assign soft weights to token representations so that one can extract highly weighted tokens as rationales \cite{7}. While this is intuitive for most of the NLP systems, the weights can be useless to give a faithful explanation because of the complex interaction of tokens. Another strand of works \cite{9,10} hard-select tokens or snippets from the input and only uses the selected part for the downstream task to get untangled explanation. This strand can be further divided into pipeline approaches and reinforcement learning approaches according to how the models are trained.
Aligning the explanation with human annotators is not necessarily the optimal objective for improving model accuracy, the various loss strategies are proposed \cite{10}. By e.g. masking out important explanation features of existing samples \cite{8}, one can augment the data. Liu et al. \cite{15} propose global supervision by adding feature attribution prior to the total loss.

\subsubsection{Generative model to generate the explanation}
% EGL using the generative model to generate the explanation}: refer to \cite{12, 16, 17, 19, 22, 23, 26, 27, 30, 31}
In addition to giving explanations directly by the attention mechanism of NLP systems, a lot of works apply additional generative models to generate natural language explanations \cite{27,30,31}. Although the rationales acquired from attention mechanisms provide concise and quick explanations, they may not have the means to provide important details of the reasoning of a model. By using an additional natural language decoder, one can generate a comprehensive description of the decision-making process behind a prediction, some examples of the generative module include a conditional random field (CRF) \cite{12},  a natural language decoder \cite{16,17}, a GRU following an MLP \cite{19}, BiLSTM and Transformer \cite{26}. The commonly used datasets and corresponding tasks are ComVE \cite{d1} for commonsense validation, e-SNLI \cite{17} for natural language inference, COSe  \cite{d2} for commonsense question answering, e-SNLI-VE \cite{d3} for visual entailment, VCR  \cite{zellers2019recognition} for visual commonsense reasoning. The ground truth is usually also natural language explanations provided by humans. To evaluate the quality of natural language explanations, one can either use automatic metrics like METEOR \cite{d21}, BERTScore \cite{d22}, and BLEURT \cite{d23},  or use human evaluation with metrics like e-ViL score \cite{d3}, confidence, and readability. In terms of the faithfulness of the natural language explanations, Wiegreffe et al. \cite{wiegreffe2020measuring} provides two necessary conditions: feature importance agreement and robustness equivalence.

\subsection{Visual Question Answering (VQA)}
% Give a summarization. Related works that can be used in this subsection: \cite{70, 71, 72, 73, 74, 75}
Attention and reasoning are two intertwined mechanisms underlying visual question-answering (VQA) tasks. Thanks to the widely used attention mechanism in VQA, applying EGL to help improve both the interpretability and performance of VQA tasks have become a hot and attractive research area in recent years. The typical EGL technique used in VQA tasks is local explanation supervision and regularization, where the sample-level visual explanation of the model is jointly optimized together with the conventional prediction loss. When applying the explanation supervision techniques, the ground truth explanation labels can be collected from human annotators\cite{70, 71, 72}, or generated by another model\cite{73, 74}. There have been a lot of VQA datasets with annotations, some are annotated with human-generated questions and answers like MovieQA \cite{tapaswi2016movieqa} and VQA v1.0 dataset in \cite{antol2015vqa}, while others are developed with synthetic scenes and rule-based templates like GQA \cite{hudson2019gqa}, Clevr \cite{johnson2017clevr}, and VCR \cite{zellers2019recognition}. VQA-2.0 \cite{goyal2017making} includes complementary images that lead to different answers, reducing language bias by forcing the model to use
visual information. The AiR-D \cite{70} is the first dataset of eye-tracking data collected from humans performing the VQA tasks. The VQA-HAT dataset \cite{das2017human} is a visual explanation dataset that
collects human attention maps by giving human experts blurred images and asking them to determine where to deblur in order to answer a given visual question. VQA-CP \cite{agrawal2018don} contains QA pairs whose distribution is significantly different between the training and test set. VQA-X \cite{park2018multimodal} offers human textual explanations which can be used to determine important objects and then are grounded to important regions in the image as the explanation. The additional attention loss can be attention accuracy\cite{70}, false sensitivity rate \cite{71} rank correlation loss \cite{73,74} and IoU loss \cite{72}. 

\subsection{Healthcare}
EGL techniques have also been well-explored in general healthcare applications, such as on gene interaction graph~\cite{greene2015understanding}, Adult Changes in Thought (ACT)~\cite{miller2017}, Mount Sinai Brain Bank (MSBB), Religious Orders Study/Memory and Aging Project (ROSMAP)~\cite{a2012overview}, and healthcare mortality prediction~\cite{miller1973plan}.
Specifically, Erion et al.~\cite{49} studied the tissue-specific gene interaction graph for the tissue most closely related to acute myeloid leukemia (AML, a type of blood cancer) in the HumanBase database~\cite{greene2015understanding} on how penalizing differences between the attributions of neighbors in an arbitrary graph connecting the features can be used to incorporate prior biological knowledge about the relationships between genes, yield more biologically plausible explanations of drug response predictions, and improve test error. They tested the model performance on a healthcare mortality prediction dataset~\cite{miller1973plan}, where the model inputs are 35 features representing patients' demographic information and medical data. 
Erion et al.~\cite{43} then further extended their previous study and proposed to add a graph attribution prior regularization on explanation to a two-layer neural network.
Their experimental results show the proposed method can significantly outperform all other methods. In addition, Weinberger et al.~\cite{53} extracted prior information from multiple gene expression datasets of the Accelerating Medicines Partnership Alzheimer’s Disease Project (AMP-AD) portal, incorporated meta-features in a gene-gene interaction graph and proposed a deep attribution prior framework to Alzheimer’s disease biomarker prediction.

\subsection{Chemistry}
EGL has also started to see emerging applications in the chemistry domain, especially for molecular puppetry prediction tasks~\cite{martins2012bayesian, subramanian2016computational, mayr2016deeptox}.
For instance, one recent work proposed an EGL framework for Graph Neural Networks (GNNs) by supervising their node- and edge-level explanation to align with domain expert annotation labels~\cite{gao2021gnes}. In this work, the authors studied three binary classification molecular datasets\footnote{Available online at: http://moleculenet.ai/datasets-1}, namely 1) The Blood-brain barrier penetration (BBBP) dataset comes from a recent study \cite{martins2012bayesian} on the modeling and prediction of barrier permeability, 2) the BACE dataset provides quantitative (IC50) and qualitative (binary label) binding results for a set of inhibitors of human b-secretase 1 (BACE-1) \cite{subramanian2016computational}, and the ``Toxicology in the 21st Century'' (TOX21) initiative created a public database measuring the toxicity of compounds \cite{wu2018moleculenet}. The general goal for each dataset is identifying functional groups on organic molecules for biological molecular properties. Each dataset contains binary classifications of small organic molecules as determined by the experiment. 
The experimental results suggest that the proposed GNES framework can effectively improve the reasonability of the explanation while still keeping or even improving the backbone GNNs model performance.

\subsection{Crime}
EGL has also been studied in the application of risk and crime-related applications, where it is important to check if the model is leveraging reasonable features when predicting crime incidences or assessing the future risk of crime suspects.
For instance, several works have studied the Propublica’s COMPAS Recidivism Risk Score datasets\footnote{Available online at: github.com/propublica/compas-analysis/}, which contains data for predicting recidivism (i.e whether a person commits a crime / a violent crime within 2 years) from many attributes~\cite{42, 62}. 
COMPAS dataset is designed for checking whether there exist biases in the mode explanation, such as the model’s treatment of the person's race attribute when making the prediction.
Specifically, Rieger et al.~\cite{42} proposed contextual decomposition explanation penalization (CDEP), a method that enables practitioners to leverage explanations to improve the performance of a deep learning model. In particular, CDEP enables inserting domain knowledge into a model to ignore spurious correlations, and correct errors, and demonstrates the ability to increase performance on real datasets; 
Alvarez et al.~\cite{62} proposed an EGL framework by explicitly enforcing three basic desiderata for interpretability—explicitness, faithfulness, and stability—during training to enhance the robustness and interpretability of model explanations. 
Besides, Balayan et al.~\cite{80} studied a private online retailer fraud detection dataset with the proposed JOEL framework, a neural network-based framework to jointly learn a decision-making task and associated explanations that convey domain knowledge. Specifically, JOEL is tailored to human-in-the-loop domain experts that lack deep technical ML knowledge, providing high-level insights about the model’s predictions that very much resemble the experts’ own reasoning. Moreover, they collect the domain feedback from a pool of certified experts and use it to ameliorate the model (human teaching), hence promoting seamless and better-suited explanations.

\subsection{Potential Future Domains of Applications}
Despite the recent attention and major advance of EGL in the aforementioned popular application domains, there are still a number of open problems and potentially fruitful directions for future research and application of EGL, as follows:

\subsubsection{FaccT}
Fairness, Accountability, and Transparency (FaccT) are becoming as important as--or depending on application areas--more important than model accuracy as an evaluation metric. 
Since it is nearly not feasible to prepare an impeccable dataset that can equally represent every possible feature related to a model's task, blindly pursuing a model's accuracy cannot exclude the chance of causing ``catastrophic consequences'' in critical circumstances~\cite{hong2020human}.
One of EGL's crucial application areas is to realize the balance between the model accuracy and FaccT by allowing human users to elicit their perspectives on steering the model.
In shaping the balance, one crucial research direction is to understand how to maximize the case where reasonable human reasoning can also cause accurate prediction.
There are several arguments discussing when human reasoning can cause a beneficial or detrimental effect on model prediction. 
While the debate is ongoing, we are gradually seeing more evidence where human involvement can result in a positive effect~\cite{gil2019towards, collaris2020explainexplore}. 
For example, Shao et al. find humans ``arguing against'' unreasonable explanation can benefit the model~\cite{shao2020towards}. 
At the end of the day, from the perspective of model accuracy and FaccT, a railroad should not the reason for predicting a train~\cite{lee2021railroad}, a snowboard cannot be a male class~\cite{hendricks2018women}, and a shopping cart should not only belong to a woman class~\cite{zhao2017men}.

\subsubsection{Adversarial Learning}
Adversarial perturbations can significantly drop the model's accuracy. In a dramatic situation, it can reach nearly to 0\%. 
Current ML models are vulnerable to adversarial attacks.
Since the majority of adversarial attack shift model's attention, applying EGL in detecting unusual shifts could be one of the solutions for developing a more robust ML model against adversarial attacks. 
However, in pursuing such a direction, the change of the attention map after the attack can be subtle from human eyes~\cite{boopathy2020proper}.
In order to apply EGL in the area of adversarial learning, we see devising better solutions in the following areas to be crucial. 
First, providing additional signals other than model attention can help human users effortlessly detect the attacked cases.
Second, devising an advanced EGL mechanism that can (1) guide the users to generate effective input (2) and applying such input to improve the model's robustness would be essential.
Following this line of thought, very recently, Jeong et al.~\cite{jeong2022learning} proposed Generative Noise Injector for Model Explanations (GNIME), a novel defense framework that perturbs model explanations to minimize the risk of model inversion attacks while preserving the interpretabilities of the generated explanations. Thus, we believe future studies on model explanation defense and attack can be one of the key research sub-areas of EGL domain.

\subsubsection{Continual \& Active Learning}
EGL's core principle is motivating ML engineers' iterative training, such as continual learning~\cite{59, 60} and active learning~\cite{21, 22}; helping them to figure out the vulnerability through explanation and fixing the issue by providing a human-level guideline.
In supporting such an iterative training, we believe one of the promising areas is ``data iteration'', a design that can help ML engineers to fortify the dataset by adding more examples based on detected vulnerabilities through explanation.
In such a direction, we believe understanding the pros and cons of retraining and continual learning can be crucial.
For example, there can be a case where newly found data points can be stacked up on an existing dataset and be used in retraining. 
Another case can be to iteratively update the last model through some of the existing techniques in continual learning~\cite{parisi2019continual}.
In general, in the world of EGL, understanding when to apply retraining or continual learning and what are the pros and cons of each training strategy are not well understood. 
Understanding which strategy can yield what strengths and weaknesses in the scenario of data iteration would be one of the core future applications of EGL.

\subsubsection{Contrastive Learning}
Contrastive learning is a powerful self-supervised learning strategy that encourages augmentations of the same input to have more similar representations compared to augmentations of different inputs.
In the field of EGL, we have started to see several works that apply the contrastive objective to the model explanation between similar/dissimilar samples to build up the explanation objective~\cite{8, 25, 63, 83}. 
The most significant advantage of leveraging the contrastive learning paradigm for explanation guidance is that no ground truth explanation annotation labels are required for model training.
However, designing an appropriate contrastive framework for EGL can be more challenging due to the lack of a standard form of model explanation under different application domains.
Besides, how to define and formulate the positive and negative explanation samples to contrast with the anchor sample's explanation can be challenging without knowing the ground-truth labels.
Thus, we believe the further development of the contrastive EGL framework can be one of the core future directions in EGL, and it can lead to a significant leap in the application of EGL to the domains where ground truth explanation labels are generally difficult to obtain in large scale.

\section{Experiments}

This section aims at providing an extensive and comprehensive experimental study among existing EGL models in various popular application domains. Specifically, the comparative studies of four datasets from the Computer Vision (CV) domain, namely 1) Gender Classification, 
2) Scene Recognition, 3) Face Glasses Recognition, and 4) Prohibited Item Detection, and three datasets from the Natural Language Processing (NLP), namely 1) Movie Review, 2) MultiRC, and 3) FEVER are provided. The details about each dataset are included in Table~\ref{tab:dataset}, where a full list of publicly available datasets for EGL is provided.

\begin{table*}[!t]
    %\centering
    \caption{A list of publicly available datasets for EGL with human annotation labels.}
    \label{tab:dataset}
     \resizebox{0.9\textwidth}{!}{
\begin{tabular}{l  c  c c }

\toprule			
Dataset & Type   & Link  & Annotation Type   \\

% vision
\midrule
Gender Classification & Vision & \url{https://github.com/YuyangGao/RES} & Pixel level \\
Scene Recognition & Vision & \url{https://github.com/YuyangGao/RES} &  Pixel level\\
Face Glasses Recognition & Vision & \url{https://github.com/carriegu0818/EGL_benchmark} &  Pixel level\\
Prohibited Item Detection & Vision & \url{https://github.com/carriegu0818/EGL_benchmark} &  Pixel level\\
ACT-X  & Vision  & \url{https://github.com/Seth-Park/MultimodalExplanations} & Pixel level and Textual  \\
Caltech-UCSD Birds  & Vision  & \url{https://authors.library.caltech.edu/27452/} & Pixel level(bounding box)  \\
The PASCAL VOC Challenge 2007  & Vision  & \url{http://host.robots.ox.ac.uk/pascal/VOC/voc2007/} & Pixel level \\
The PASCAL VOC Challenge 2012  & Vision  & \url{http://host.robots.ox.ac.uk/pascal/VOC/voc2012/} & Pixel level \\
ISIC2018 Challenge  & Vision  & \url{https://challenge.isic-archive.com/landing/2018/} & Pixel level(bounding box)  \\
Pneumonia Detection  & Vision  & \url{https://www.kaggle.com/c/rsna-pneumonia-detection-challenge} & Pixel level(bounding box)  \\

% NLP
\midrule
Movie Review & NLP & \url{https://github.com/jayded/eraserbenchmark} & Span–level rationale\\
MultiRC & NLP & \url{https://github.com/jayded/eraserbenchmark} & Single sentence-level rationale \\
FEVER & NLP & \url{https://github.com/jayded/eraserbenchmark} & Sentence-level rationale  \\
BoolQ & NLP & \url{https://github.com/jayded/eraserbenchmark} & Token-level rationale  \\
Evidence inference  & NLP & \url{https://github.com/jayded/eraserbenchmark} & Sentence-level rationale  \\
e-SNLI  & NLP & \url{https://github.com/jayded/eraserbenchmark} & Token-level rationale  \\
Commonsense Explanations (CoS-E)  & NLP & \url{https://github.com/jayded/eraserbenchmark} & Sentence-level rationale  \\

\midrule
% VQA
VQA-HAT  & VQA & \url{https://computing.ece.vt.edu/~abhshkdz/vqa-hat/} & Pixel level   \\
GQA  & VQA  & \url{https://cs.stanford.edu/people/dorarad/gqa/about.html} & Pixel level   \\
VQA-X  & VQA  & \url{https://github.com/Seth-Park/MultimodalExplanations} & Pixel level  and Textual  \\
VQS  & VQA  & \url{https://github.com/Cold-Winter/vqs} & Pixel level  \\

% Graph
\midrule
BBBP & Graph & \url{https://github.com/YuyangGao/GNES} & Node- and edge–level\\
BACE & Graph & \url{https://github.com/YuyangGao/GNES} & Node- and edge–level \\
TOX21 & Graph & \url{https://github.com/YuyangGao/GNES} & Node- and edge–level  \\

% others (if applicable)
% ... 

\bottomrule
\end{tabular}
}
\end{table*} 

%table, dataset, category, citation, link, annotation type
%delete footnote
%github link for own dataset - Explanation Guided Learning Benchmark

\subsection{Visual Explanation Guided Learning}

\subsubsection{Gender Classification~\cite{gao2022res}} The gender classification task is derived from the Microsoft COCO dataset~\footnote{Available at: \url{https://cocodataset.org/}}~\cite{lin2014microsoft} by extracting images that had the word ``men'' or ``women'' in their captions. 
The dataset is further filtered by removing images with 1) both gender in the caption, 2) multiple people present, or 3) not recognizable humans. A subset of the images is further manually annotated by human annotators as factual and counterfactual masks.
The dataset in total consists of 1,736 images with human annotations, where the distribution of female to male is even. 
For data splitting, we only randomly sampled 100 samples as the training set to better simulate a more practical situation where we only have limited access to the human explanation labels.
The validation and test set is set to 700.

\subsubsection{Scene Recognition Dataset~\cite{gao2022res}}
The scene recognition dataset is originally derived from the Places365 dataset\footnote{Available at: \url{http://places2.csail.mit.edu/index.html}}~\cite{zhou2017places} and manually annotated by Gao et al.~\cite{gao2022res}. The task for this dataset is a binary classification of scene recognition: nature vs. urban. 
Specifically, the categories used to sample the data are listed below:
\begin{itemize}
    \item \textit{Nature}: mountain, pond, waterfall, field wild, forest broadleaf, rainforest
    \item \textit{Urban}: house, bridge, campus, tower, street, driveway
\end{itemize}
The dataset consists of a total of 2086 images with human explanation labels. Similarly, we split the data randomly with a sample size of 100/700/700 for training, validation, and testing.

\subsubsection{Face Glasses Recognition} We construct the glasses recognition dataset from the CelebAMask-HQ dataset~\footnote{Available at: \url{http://mmlab.ie.cuhk.edu.hk/projects/CelebA/CelebAMask_HQ.html}}~\cite{CelebAMask-HQ} by categorizing face images with and without glasses.
In CelebAMask-HQ, masks were manually annotated with 19 classes including all facial components and accessories.
The rationale of the task is that we are able to obtain factual annotation labels by the segmentation of eyes and glasses directly.
While the original dataset is highly imbalanced in the ratio between faces with and without glasses, we randomly select an equal number of images in both classes, with a total of 100/393/392 images for training/validation/testing respectively.\\
\subsubsection{Prohibited Item Detection} The task is constructed from the Sixray dataset \footnote{Available online at: https://github.com/MeioJane/SIXray}~\cite{Miao2019SIXray} by splitting images based on the presence of prohibited items.
Sixray is highly imbalanced with 1,059,231 X-ray images, including 6 classes of 8,929 prohibited items. Merging the 6 prohibited classes, the task of the new dataset is a binary prohibited item detection. Bounding boxes of prohibited items are included in all images. Due to data imbalance, the dataset is further filtered into 100/5296/5298 images for training, validation, and testing respectively.

\subsubsection{Evaluation Metrics}
We evaluate the model in terms of prediction performance as well as in terms of explanation performance.
For prediction performance, we use AUC and accuracy as evaluation metrics.
To evaluate explanation faithfulness, we employ the Matrix for comprehensiveness and sufficiency by ERASER~\cite{7}. 
For explanation correctness assessment, we compare the saliency map generated by Grad-CAM with ground-truth annotation masks. 
Specifically, we use the Intersection over Union (IoU) score~\cite{bau2017network},  the bit-wise intersection and union operations between the ground truth explanation and the binarized model explanation. We further evaluate explanation performance with Explanatory F1, precision, and recall by bit-wise comparison between ground-truth explanation and model explanation.

\subsubsection{Comparison methods}:
We compare the performance of several models as listed below:

\begin{itemize}
    \item \textbf{Baseline} Baseline 1 and 2 are pre-trained ResNet50 and VGG16 model that trains only on prediction loss without explanation loss. 
    \item \textbf{GRADIA}~\cite{gao2022aligning}: A framework that trains the DNN model with both the prediction loss as well as a conventional L1 loss that directly minimizes the distance between the continuous model explanation and the binary positive explanation labels.
    \item \textbf{RES}~\cite{gao2022res}: A framework that trains the DNN with both factual and counterfactual annotations with two imputation functions: $g(\cdot)$ as a fixed value Gaussian convolution filter and learnable imputation function $g_\phi(\cdot)$ via multiple layers of learnable kernels.
    \item \textbf{CDEP}~\cite{42}: A framework that incorporates Contextual Decomposition (CD) to penalize spurious correlations and therefore correct errors.
    \item \textbf{RRR}~\cite{5}: A framework that was initially introduced by~\cite{4} and altered by~\cite{5}, which aims to regularize the model to be right for the right reasons.
    \item \textbf{SGT}~\cite{68}: A framework that introduces saliency-guided training for neural networks to reduce noisy gradients in predictions.
    \item \textbf{SENN}~\cite{62}: A framework that applied two regularization terms on the model explanation: 1) neighborhood-fidelity and 2) stability based on the neighborhood of input.
\end{itemize}

\subsubsection{Implementation Details}:
All models are trained for 50 epochs with the same train/val/test split as mentioned above. We use the ADAM optimizer with a learning rate of 0.0001~\cite{kingma2014adam}. The architecture of each model is listed in Table \ref{tab:gender}. To better compare the performance on explainability, the model explanations are generated by Grad-CAM~\cite{selvaraju2017grad}. When calculating the explanation evaluation metrics, the explanation maps were further binarized by a fixed threshold of $0.5$. We use a batch size of 32 for training and 100 for testing. For GRADIA and RES, we set the slack variable $\alpha$ to 0.1 and 0.01, respectively, and the regularization factor to 0. For RRR, we set the regularization parameter to 1.  For CDEP, we set the regularizer rate to 0, 0.1, and 10. For SENN, we set the robust regularization, sparsity regularization, and concept regularization hyperparameters to 0.0001, 0.00002, and 1, respectively. For SGT, we set features dropped to 0.1 and 0.3.
\\

\subsubsection{Quantitative analysis
}
 \begin{table*}%[!t]
    %\centering
    \caption{The classification performance and explanation evaluation on Gender Classification. The results are obtained from 3 individual runs and the best results of each metric are highlighted in boldface font. }
    \label{tab:gender}
     \resizebox{0.9\textwidth}{!}{
\begin{tabular}{l | c | c c | c c | c c c c}
\toprule
 \multicolumn{2}{c}{} & \multicolumn{2}{c}{Prediction} & \multicolumn{2}{c}{Exp Faithfulness}  & \multicolumn{4}{c}{Exp Correctness} \\

\midrule
			
Model & Architecture & Acc. $\uparrow$  & AUC $\uparrow$ & Comp. $\uparrow$ & Suff. $\downarrow$ & IoU  $\uparrow$ & F1 $\uparrow$ &  Precision $\uparrow$ &  Recall $\uparrow$\\
\midrule
Baseline 1 & ResNet50 & 0.680 & 0.664 & 0.048 & 0.105 & 0.147 & 0.470 & 0.554 & 0.525 \\

Baseline 2 & VGG16 & 0.637 & 0.671 & 0.048 & 0.051 & 0.058 & 0.155 & 0.340  & 0.132 \\
\midrule
\multicolumn{10}{c}{Supervised}\\
\midrule
GRADIA\cite{gao2022aligning} & ResNet50 & \textbf{0.695} & \textbf{0.764} & 0.107 & 0.090 & \textbf{0.243} & \textbf{0.625} & \textbf{0.787} & 0.608 \\
\midrule
RES~\cite{gao2022res}        & ResNet50 & 0.690 & 0.744 & \textbf{0.108} & 0.097 & 0.240 & 0.614 & 0.742 & \textbf{0.621} \\
\midrule
%RRR & MLP & 0.533 & 0.558 &  &  &  &  \\
%\midrule

RRR\cite{5} & VGG16 & 0.624 & 0.628 & 0.027 & 0.015 & 0.091 & 0.270 & 0.424 & 0.257 \\
\midrule
CDEP~\cite{42}        & VGG16 & 0.628 &  0.635 & 0.014 & 0.010  & 0.100 &  0.254 & 0.419 & 0.231 \\
\midrule
\multicolumn{10}{c}{Unsupervised}\\
\midrule
SENN~\cite{62}        & SENN(CNN) & 0.589 &  0.627 & 0.005 & 0.027 & 0.062 & 0.186 & 0.300 & 0.186 \\

\midrule
SGT~\cite{68}       & ResNet50 & 0.645 &  0.687 & 0.012 & \textbf{0.002} & 0.044 & 0.352 & 0.502 & 0.257 \\

\bottomrule
\end{tabular}
}
\end{table*} 

\begin{table*}%[!t]
    %\centering
    \caption{The classification performance and explanation evaluation on Scene Recognition. The results are obtained from 3 individual runs and the best results of each metric are highlighted in boldface font. }
    \label{tab:scene}
     \resizebox{0.9\textwidth}{!}{
\begin{tabular}{l | c | c c | c c | c c c c}
\toprule
 \multicolumn{2}{c}{} & \multicolumn{2}{c}{Prediction} & \multicolumn{2}{c}{Exp Faithfulness}  & \multicolumn{4}{c}{Exp Correctness} \\

\midrule
			
Model & Architecture & Acc. $\uparrow$  & AUC $\uparrow$ & Comp. $\uparrow$ & Suff. $\downarrow$ & IoU  $\uparrow$ & F1 $\uparrow$ &  Precision $\uparrow$ &  Recall $\uparrow$\\

\midrule
Baseline 1 & ResNet50 & 0.947 & 0.965 & 0.068 & 0.255 & 0.397 & 0.702 & 0.906  & 0.628 \\

Baseline 2 & VGG16 & 0.953 & \textbf{0.988} & 0.134 & 0.117 & 0.191 & 0.324 & 0.890  & 0.226 \\ 
\midrule
\multicolumn{10}{c}{Supervised}\\
\midrule
GRADIA~\cite{gao2022aligning} & ResNet50 & 0.952 & 0.987 & \textbf{0.255} & 0.073 & 0.378 & 0.606 & 0.912 & 0.501 \\
\midrule
RES~\cite{gao2022res}         & ResNet50 & \textbf{0.956} & \textbf{0.988} & 0.189 & \textbf{0.002} & \textbf{0.435} & \textbf{0.722} & \textbf{0.909} & \textbf{0.647} \\
\midrule
%RRR & MLP & 0.643 & 0.704 &  &  &  &  \\
%\midrule

RRR~\cite{5} & VGG16 & 0.953 &  0.987 & 0.014 & 0.019 & 0.224 & 0.364 & 0.925 & 0.250 \\
\midrule
CDEP~\cite{68}       & VGG16 & 0.934 & 0.952 & 0.026 & 0.039 & 0.127 & 0.232 & 0.807 & 0.153  \\
\midrule
\multicolumn{10}{c}{Unsupervised}\\
\midrule
SENN~\cite{42}      & SENN(CNN) & 0.733  &  0.798 & 0.022 & 0.042 & 0.082 & 0.183 & 0.721 & 0.108 \\

\midrule
SGT~\cite{62}       & ResNet50 & 0.937 & 0.985 & 0.164 & 0.039 & 0.056 & 0.301 & 0.796 & 0.213 \\

%\midrule
%CALPI         & Di-SENN(MLP) &  &   &  &  &  &  \\

\bottomrule
\end{tabular}
}
\end{table*} 

\begin{table*}[!t]
    %\centering
    \caption{The classification performance and explanation evaluation on the Face Glasses Recognition. The results are obtained from 3 individual runs and the best results of each metric are highlighted in boldface font.}
    \label{tab:face}
     \resizebox{0.9\textwidth}{!}{
\begin{tabular}{l | c | c c | c c | c c c c}
\toprule
 \multicolumn{2}{c}{} & \multicolumn{2}{c}{Prediction} & \multicolumn{2}{c}{Exp Faithfulness}  & \multicolumn{4}{c}{Exp Correctness} \\

\midrule
			
Model & Architecture & Acc. $\uparrow$  & AUC $\uparrow$ & Comp. $\uparrow$ & Suff. $\downarrow$ & IoU  $\uparrow$ & F1 $\uparrow$ &  Precision $\uparrow$ &  Recall $\uparrow$\\

\midrule
Baseline 1 & ResNet50 & 0.991 & \textbf{0.999} & 0.302 & 0.163 & 0.134 & 0.971 & 0.998  & 0.954 \\

Baseline 2 & VGG16 &\textbf{ 0.996} & 0.864 & 0.183 & 0.039 & 0.299 & 0.873 & 0.987  & 0.804 \\
\midrule
\multicolumn{10}{c}{Supervised}\\
\midrule

GRADIA~\cite{gao2022aligning} & ResNet50 & 0.990 & \textbf{0.999} & 0.368 & 0.262 & \textbf{0.375} & \textbf{0.949} & 0.993 & \textbf{0.917} \\
\midrule
RES~\cite{gao2022res}   & ResNet50 & 0.991 & \textbf{0.999} & \textbf{0.396} & 0.128 & 0.302 & 0.932 & \textbf{0.997} & 0.887 \\
\midrule

RRR~\cite{5} & VGG16 & 0.994 & \textbf{0.999} & 0.384 & 0.160 & 0.332 & 0.909 & 0.992 & 0.864 \\
\midrule
CDEP~\cite{42}         & VGG16 & \textbf{0.996} & \textbf{0.999}  & 0.004 & \textbf{0.003} & 0.042 & 0.203 & 0.540 & 0.248 \\

\midrule
\multicolumn{10}{c}{Unsupervised}\\
\midrule
SENN~\cite{62}      & SENN(CNN) & 0.797 &  0.873 & 0.002 & 0.023 &0.045 & 0.202 & 0.639 &  0.134 \\

\midrule
SGT~\cite{68}         & ResNet50 & \textbf{0.996} &  \textbf{0.999} & 0.083 & 0.106 & 0.292 & 0.671 & 0.974 & 0.556 \\

\bottomrule
\end{tabular}
}
\end{table*} 

\begin{table*}[!t]
    %\centering
    \caption{The classification performance and explanation evaluation on the Prohibited Item Detection. The results are obtained from 3 individual runs and the best results of each metric are highlighted in boldface font.}
    \label{tab:sixray}
     \resizebox{0.9\textwidth}{!}{
\begin{tabular}{l | c | c c | c c | c c c c}
\toprule
 \multicolumn{2}{c}{} & \multicolumn{2}{c}{Prediction} & \multicolumn{2}{c}{Exp Faithfulness}  & \multicolumn{4}{c}{Exp Correctness} \\

\midrule
			
Model & Architecture & Acc. $\uparrow$  & AUC $\uparrow$ & Comp. $\uparrow$ & Suff. $\downarrow$ & IoU  $\uparrow$ & F1 $\uparrow$ &  Precision $\uparrow$ &  Recall $\uparrow$\\

\midrule

Baseline 1 & ResNet50 & 0.961 & 0.992 & 0.161 & 0.053 & 0.195 & 0.823 & 0.870  & 0.784 \\
Baseline 2 & VGG16 & 0.917 & 0.988 & -0.026 & \textbf{-0.010} & 0.147 & 0.330 & 0.788  & 0.241 \\

\midrule
\multicolumn{10}{c}{Supervised} \\
\midrule
GRADIA~\cite{gao2022aligning} & ResNet50 & \textbf{0.974} & \textbf{0.997} & \textbf{0.176} & 0.272 & 0.213 & 0.703 & 0.928 & 0.610 \\
\midrule
RES~\cite{gao2022res}         & ResNet50 & 0.962 & 0.995 & 0.152 & 0.295 & \textbf{0.235} & \textbf{0.837} & \textbf{0.964} & \textbf{0.776} \\
\midrule
%RRR & MLP & 0.775 & 0.911 &  &  &  &  \\
%\midrule
RRR~\cite{5} & VGG16 & 0.950 & 0.995 &  0.055 & 0.128 & 0.155  & 0.343 & 0.448 & 0.320  \\
\midrule
CDEP~\cite{42}         & VGG16 & 0.959 & 0.992 & 0.038 & 0.021 & 0.061 & 0.403 & 0.615 & 0.298 \\
\midrule
\multicolumn{10}{c}{Unsupervised} \\
\midrule
SENN~\cite{62}  & SENN(CNN) & 0.754 &  0.839 & -0.026 & 0.095 & 0.042 & 0.152 &  0.584& 0.098 \\

\midrule
SGT~\cite{68}         & ResNet50 & 0.962 & 0.993 & 0.027 & 0.066 & 0.062 & 0.552 & 0.648 & 0.481 \\

%\midrule
%CALPI         & Di-SENN(MLP) &  &   &  &  &  &  \\

\bottomrule
\end{tabular}
}
\end{table*} 

Model prediction performance and explanation quality in the domain of computer vision are presented in Tables \ref{tab:gender}, \ref{tab:scene}, \ref{tab:face}, and \ref{tab:sixray}. We evaluate 6 models through gender classification, scene recognition, glasses identification, and prohibited item discovery. We evaluate two baseline model performances: ResNet50 and VGG16, as they are employed by the selected paper. Overall, supervised models demonstrate better prediction power and higher explanation quality than unsupervised models. ResNet50 seems to perform slightly better in prediction, and significantly better in explanation quality compared with VGG16.

For gender classification, GRADIA generally has the best prediction performance and presents the highest explanation quality, with the best scores in all metrics besides comprehensiveness and Exp recall, and minimal differences of 0.9\& and 2.1\% in comprehensiveness and explanatory recall. For models with ResNet50 as the backbone architecture, GRADIA and RES significantly outperform SGT, since SGT is unsupervised. SGT presents a lower accuracy, comprehensiveness, IoU, and explanatory F1 than the baseline model, which implies that the unsupervised model is not improving model performance and explanation quality. Yet SGT achieves the lowest sufficiency, which measures how well the prediction aligns between the original input and an explanation-generated input. Models with VGG16 as the backbone report lower sufficiency than those with ResNet50, while models with ResNet50 generally hold higher accuracy, comprehensiveness, IoU, and explanatory F1. SENN, which develops its own architecture with a set of conceptizer, parametrizer, and aggregator, underperforms in all metrics since the backbone is a simple CNN model. 

For scene recognition, the baseline VGG16 achieves better accuracy, comprehensiveness, and sufficiency, compared with the baseline ResNet50. VGG16 has a significantly low explanatory recall and IoU, which results in 53.8\% worse performance in explanatory F1. For the selected models, RES yields the best performance on all metrics, slightly improving prediction performance and boosting explanation quality significantly, with 1.0\%, 2.4\%, 177.9\%, -99.2\%, 9.6\%, 2.8\% changes in accuracy, AUC, comprehensiveness, sufficiency, IoU, and explanatory F1, respectively. SENN consistently underperforms in all metrics. Among models with VGG16, RRR is able to maintain a similar accuracy as the baseline while improving explanation correctness (IoU and Exp F1) by 17.3\% and 12.3\%, while CDEP results in worse explanation correctness. RRR and CDEP both obtain a lower score in comprehensiveness and sufficiency, which suggests that even stripping off the model-generated explanations, the models are able to generate similar predictions. The stripped model-generated explanations are useful in terms of prediction, but not all useful information is covered by the saliency map. Moreover, baseline ResNet50 under-performs in terms of explanation faithfulness but outperforms in terms of explanation correctness. This implies that while ResNet50 is able to generate explanations that exhibit the pattern of human annotation, the generated explanations are not useful for the model in terms of prediction.

In the task of the glasses identification, all models achieve high performance besides SENN, which consistently results in worse performance in all metrics. In terms of explanation comprehensiveness and faithfulness, GRADIA, RES, RRR, and CDEP show (21.9\%, 60.7\%), (31.1\%, -21.5\%), (109.8\%, 310.3\%), and (-97.8\%, -92.3\%) changes with respect to the baseline. RES holds the highest comprehensiveness score and CDEP holds the lowest sufficiency score. This implies that GRADIA, RES, and RRR are successful at extracting all useful attention for prediction, while GRADIA and RRR sacrifice the prediction power if solely using generated attention as the input. Among supervised models, GRADIA has the highest IoU as well as the highest percentage increase, which implies that the model successfully learns the pattern of human annotation and is able to produce saliency maps that are most aligned with human annotation. However, since GRADIA also has the highest sufficiency score, it further implies that learning from the human annotation may not be sufficient for the model to make correct predictions. Meanwhile, SGT shows high accuracy and IoU even as an unsupervised model. Yet it suffers from low Explanatory recall, low comprehensiveness, and high sufficiency score, indicating that the model-generated attention is not successful in terms of prediction.

In the Prohibited Item Detection task, models generally exhibit a pattern of maintaining high accuracy, better explanation correctness, and comprehensiveness, but worse sufficiency. ResNet50 baseline achieves higher prediction and explanation robustness compared with VGG16. 
While most models' accuracy ranges from 0.917 to 0.974, SENN yields an accuracy of 0.754, which implies that a CNN model is insufficient for the dataset. 
All models with VGG16 as the backbone performs well in terms of low sufficiency but poorly in comprehensiveness and explanatory recall. GRADIA yields the highest accuracy (0.974), AUC (0.997) and comprehensiveness score (0.176), but poor sufficiency score of 0.272 when the baseline has a sufficiency of 0.053. 
RES shows the worst good performance on sufficiency but second to best score on comprehensiveness among EGL models. 
The baseline model with VGG16 achieves the best sufficiency score of -0.010. This is because the model sufficiency score is highly influenced by the size of generated annotation map, and the VGG16 baseline model scarifies the sparseness of explanation to achieve a higher sufficiency and consequently leads to the worst comprehensiveness score. To validate the above statement, we further compute the proportion of attention map generated respectively to the entire image of the VGG16 baseline, RES, and GRADIA models. We find that the average explanation map sizes of the VGG16 baseline model are on average 84\% and 33\% greater than those of RES and GRADIA, respectively. This additional observation provides additional support for our assumption that the baseline model tends to generate larger explanation maps, leading to a much higher sufficiency score but a much worse comprehensiveness score.
In terms of explanation correctness, RES and RRR are able to improve IoU from 0.195 to 0.101 and from 0.147 to 0.155, the best among each architecture. RES improves explanatory F1 from 0.827 to 0.837 and CDEP improves explanatory F1 from 0.330 to 0.403. Overall, GRADIA achieves the best prediction accuracy and faithfulness while RES achieves the best explanatory correctness among all models. 

\subsubsection{Qualitative Case Study}

\begin{figure*}
    \centering
    \includegraphics[width=0.9\textwidth]{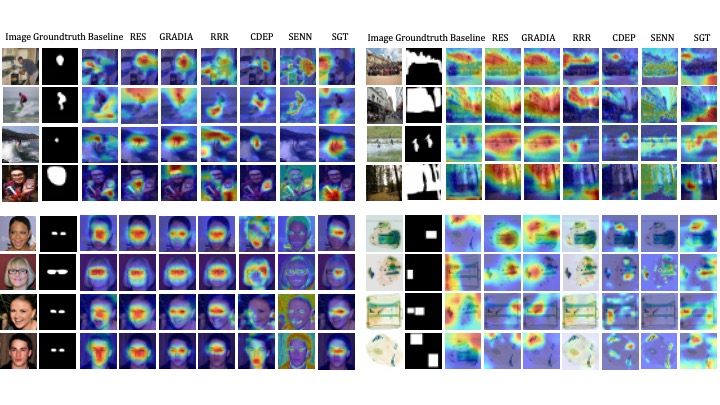}
    \caption{Selected explanation visualization results on four vision datasets: gender classification, scene recognition, face glasses detection, and prohibited item identification. The model-generated explanations are highlighted.}
    \label{fig:exp_viz4}
\end{figure*}

Figure 5 displays the visualization results of the four vision tasks: gender classification, scene recognition, face glasses detection, and prohibited item identification. Overall, RES and GRADIA have more overlap with the ground truth. CDEP is more fine-grained. SENN generates a saliency map that highlights all areas instead of focusing on a particular place. In the gender classification task, RES and GRADIA perform well in identifying the human body, even with distractions. RRR, CDEP, and SGT sometimes highlight areas that are disruptive, such as phones and kitchen stoves. In terms of saliency map size, CDEP generates a saliency map that focuses only on a small area, whereas SENN generates a thin layer of attention all over the image, which explains their low performance in IoU and explanatory F1. In the scene recognition task, RES, GRADIA, and SENN have the most overlapping areas with the ground truth label. RRR and CDEP sometimes are biased. For example, in the last image, RRR considers the floor key elements in scene recognition, whereas the ground truth label is the trees. SGT mostly attends to areas other than the ground truth, which explains its low IoU compared with other models. For face glasses detection, all models generate saliency maps focused on the face area. While the baseline model focus on the entire face, RES, GRADIA, RRR, and SGT generate maps that are more specific to the eye areas. CDEP focuses more on the lower half of the face and SENN attends to all face parts, such as the eyes, mouth, etc. While all models are highly accurate at identifying the presence of prohibited items, model-generated maps do not align with ground truth labels. Most models are able to focus on some small objects, not necessarily the prohibited items. SGT focuses on the white area outside of the baggage, which accounts for its low accuracy and low explanation performance. 

\subsection{Rationale Attention Guided Learning}
To evaluate the performance of EGL models on NLP tasks, three datasets with explanation rationales are selected for the experimental study~\cite{7}, the details about each dataset are shown as follows: 
\subsubsection{Movie Review~\cite{gao2022res}} The movie review dataset includes binary sentiment labels as well as rationale annotations at the span level. The task is to classify movies with positive sentiments from those with negative sentiments. we randomly split the data with a sample size of 1600/150/200 for training, validation, and testing. The explanation label is the sentiment $\in\{$positive, negative$\}$.
\subsubsection{MultiRC~\cite{MultiRC2018}} MultiRC is a reading comprehension dataset originally composed of a series of rationale/question/answer triplets. This is also a binary classification task where the prediction label is True or False. The dataset is divided into  24029/3214/4848 for training, validation, and testing. The ground truth for explanation indicates if the answer is correct. 
\subsubsection{FEVER~\cite{thorne-etal-2018-fever}} FEVER is a fact verification dataset, where each claim can be classified into supported, refuted, or not enough information. DeYoung et al.~\cite{7} further took a subset of the dataset and included only support and refuted claims. The dataset is further separated into 97957/6122/6111 images for training, validation, and testing respectively. For explanation rationales, the model
has to predict the veracity of a claim $\in\{$support, refuse$\}$.

\subsubsection{Evaluation Metrics}:
We evaluate the model in two categories: 1) prediction performance and 2) explanation performance.
For prediction performance, accuracy and AUC are computed to evaluate the predictive power of the model. 
For explanation evaluation, we incorporated 6 matrices to fully examine the explanation robustness. Matrix for comprehensiveness and sufficiency are derived from ERASER ~\cite{7}. In addition, we measure the token level intersection over union (IoU)~\cite{bau2017network} between ground truth rationale and predicted rationale through IoU, explanatory F1, precision, and recall. 

\subsubsection{Comparison methods}:
We compare the performance of several models listed below:
\begin{itemize}
    
    \item \textbf{Baseline}: Baselines 1, 2, and 3 are pre-trained models that train only using the prediction loss without explanation loss. The pre-trained architecture for baselines 1, 2, and 3 are BERT+MLP, BERT+LSTM, and BERT+BERT respectively.
    
    \item \textbf{ERASER}~\cite{7}: A pipeline model that first trains the encoder to extract rationales, and then trains the decoder to perform prediction using only rationales. 
    
    \item \textbf{Glockner et al.}~\cite{9}: A differentiable training–framework that aims to output faithful rationales
on a sentence level

    \item \textbf{Carton et al.}~\cite{10}: A model that applies sentence-level rationale supervision, non-occluding “importance
embeddings” on selective rationales with high sufficiency-accuracy.
    
    \item \textbf{Expred}~\cite{19}: A novel explanation generation framework work using multi-task learning that is task-aware and can exploit rationales data for effective explanations.
    
    \item \textbf{FRESH}~\cite{31}: A model that aims to produce faithful rationales for neural text classification by defining
independent snippet extraction and prediction modules.

\end{itemize}

\subsubsection{Implementation Details}: The data preprocessing follows the setting of ERASER~\cite{7}. We train all the models equally for $20$ epochs and Adam is used for optimization with a learning rate of $2e$-$5$. To evaluate the explanation performance, the threshold for the calculated rationales is set to be $0.5$. We follow the hyperparameter settings reported in the papers of the above methods.

\subsubsection{Quantitative analysis}

\begin{table*}%[!t]
    %\centering
    \caption{The classification performance and explanation evaluation on the Movie Review dataset. The best results of each metric are highlighted in boldface font.}
    \label{tab:Movie}
     \resizebox{0.9\textwidth}{!}{
\begin{tabular}{l | c | c c | c c | c c c c}
\toprule
  \multicolumn{2}{c}{} & \multicolumn{2}{c}{Prediction} & \multicolumn{2}{c}{Exp Faithfulness}  & \multicolumn{4}{c}{Exp Correctness} \\

\midrule
			
Model & Architecture & Acc. $\uparrow$  & AUC $\uparrow$ & Comp. $\uparrow$ & Suff. $\downarrow$ & IoU  $\uparrow$ & F1 $\uparrow$ &  Precision $\uparrow$ &  Recall $\uparrow$\\
\midrule
Baseline 1 & BERT+MLP  & 0.516 & 0.478 & 0.086 & 0.145 & 0.242 & 0.365 & 0.441 & 0.312 \\
\midrule
Baseline 2 & BERT+LSTM & 0.622 & 0.591 & 0.027 & 0.126 & 0.043 & 0.112 & 0.462 & 0.064 \\
\midrule
Baseline 3 & BERT+BERT & 0.756 & 0.703 & 0.112 & 0.113 & 0.085 & 0.188 & 0.411 & 0.122 \\
\midrule
ERASER \cite{7} & BERT+LSTM & 0.826 & 0.805 & 0.128 & 0.093 & 0.598 & 0.749 & 0.734 &  0.765 \\
\midrule
Glockner et al.\cite{9}    & BERT+MLP & 0.564 & 0.511 & 0.114 & 0.103 & 0.541 & 0.702 & 0.693 &  0.712\\
\midrule

Carton et al.~\cite{10} & BERT+BERT & \textbf{0.834 } & \textbf{0.812} & 0.138 & 0.084 & 0.585 & 0.738 & 0.726 &  0.751 \\

\midrule

Expred~\cite{19}   & BERT+GRU+MLP & 0.794 & 0.779 & 0.094 & \textbf{0.076} & \textbf{0.639} & \textbf{0.779} & \textbf{0.781} &  \textbf{0.779} \\

\midrule
FRESH~\cite{31}        & BERT+LSTM & 0.678 & 0.653 & \textbf{0.144} & 0.093 & 0.569 & 0.726 & 0.745 &  0.707 \\

\bottomrule
\end{tabular}
}
\end{table*} 

\begin{table*}%[!t]
    %\centering
    \caption{The classification performance and explanation evaluation on the MultiRC dataset. The best results of each metric are highlighted in boldface font.}
    \label{tab:MultiRC}
     \resizebox{0.9\textwidth}{!}{
\begin{tabular}{l | c | c c | c c | c c c c}
\toprule
 \multicolumn{2}{c}{} & \multicolumn{2}{c}{Prediction} & \multicolumn{2}{c}{Exp Faithfulness}  & \multicolumn{4}{c}{Exp Correctness} \\

\midrule
			
Model & Architecture & Acc. $\uparrow$  & AUC $\uparrow$ & Comp. $\uparrow$ & Suff. $\downarrow$ & IoU  $\uparrow$ & F1 $\uparrow$ &  Precision $\uparrow$ &  Recall $\uparrow$\\
\midrule
Baseline 1 & BERT+MLP  & 0.564 & 0.511 & 0.012 & 0.188 & 0.235 & 0.459 & 0.534 & 0.402 \\
\midrule
Baseline 2 & BERT+LSTM & 0.593 & 0.573 & 0.081 & 0.205 & 0.106 & 0.280 & 0.471 & 0.199 \\
\midrule
Baseline 3 & BERT+BERT & 0.627 & 0.580 & 0.054 & 0.154 & 0.076 & 0.234 & 0.485 & 0.154\\
\midrule
ERASER~\cite{7} & BERT+LSTM & 0.639 & 0.615 & 0.039 & 0.132 & 0.448 & 0.618 & 0.615 &  0.622 \\
\midrule
Glockner et al.~\cite{9}    & BERT+MLP & 0.587 & 0.547 & 0.065 & 0.136 & 0.409 & 0.580 & 0.576 &  0.585\\
\midrule

Carton et al.~\cite{10} & BERT+BERT & \textbf{0.647} & 0.613 & 0.074 & 
0.076 & \textbf{0.473} & \textbf{0.642} & \textbf{0.633} &  \textbf{0.651} \\

\midrule

Expred~\cite{19}   & BERT+GRU+MLP & 0.638 & \textbf{0.622} & 0.032 & \textbf{0.061} & 0.447 & 0.618 & 0.602 &  0.635 \\

\midrule
FRESH~\cite{31}         & BERT+LSTM & 0.607 & 0.586 & \textbf{0.096} & 0.113 & 0.437 & 0.608 & 0.613 &  0.604 \\

\bottomrule
\end{tabular}
}
\end{table*} 

\begin{table*}%[!t]
    %\centering
    \caption{The classification performance and explanation evaluation on the Fever dataset. The best results of each metric are highlighted in boldface font.}
    \label{tab:fever}
    \resizebox{0.9\textwidth}{!}{
\begin{tabular}{l | c | c c | c c | c c c c}
\toprule
 \multicolumn{2}{c}{} & \multicolumn{2}{c}{Prediction} & \multicolumn{2}{c}{Exp Faithfulness}  & \multicolumn{4}{c}{Exp Correctness} \\

\midrule
			
Model & Architecture & Acc. $\uparrow$  & AUC $\uparrow$ & Comp. $\uparrow$ & Suff. $\downarrow$ & IoU  $\uparrow$ & F1 $\uparrow$ &  Precision $\uparrow$ &  Recall $\uparrow$\\
\midrule
Baseline 1 & BERT+MLP  & 0.822 & 0.803 & 0.075 & 0.126 & 0.103 & 0.319 & 0.513 & 0.231 \\
\midrule
Baseline 2 & BERT+LSTM & 0.851 & 0.822 & 0.022 & 0.099 & 0.157 & 0.391 & 0.454 & 0.344 \\
\midrule
Baseline 3 & BERT+BERT & 0.872 & 0.856 & 0.017 & 0.117 & 0.036 & 0.145 & 0.612 & 0.082 \\
\midrule
ERASER~\cite{7} & BERT+LSTM & 0.874 & 0.867 & 0.036 & 0.053 & 0.679 & 0.808 & 0.805 &  0.812 \\
\midrule
Glockner et al.~\cite{9}    & BERT+MLP & 0.835 & 0.813 & \textbf{0.122} & 0.066 & 0.672 & 0.803 & \textbf{0.833 } &  0.776\\
\midrule

Carton et al.~\cite{10} & BERT+BERT & 0.893 & 0.876 & 0.084 & 0.048 &\textbf{ 0.707} & \textbf{0.828} & 0.831 &  \textbf{0.826} \\

\midrule

Expred~\cite{19}   & BERT+GRU+MLP & \textbf{0.903} & \textbf{0.889} & 0.043 & \textbf{0.027} & 0.696 & 0.820 & 0.817 &  0.824 \\

\midrule
FRESH~\cite{31}         & BERT+LSTM & 0.862 & 0.832 & 0.106 & 0.053 & 0.627 & 0.771 & 0.732 &  0.814 \\

\bottomrule
\end{tabular}
}
\end{table*} 

Tables 6, 7, and 8 present the model prediction performance and explanation quality of Movie Review, MultiRC, FEVER dataset respectively. The best results for each dataset are highlighted with boldface font. In general, when comparing with baseline, all models achieve a better accuracy and explanation correctness. The sufficiency score also decreases compared with the baseline model, which implies that the model-generated rationale is representative of the entire document. 

For the Movie Review dataset, Carton et al.~\cite{10} yields the highest classification accuracy and Expred~\cite{19} generates the explanations with the highest quality. Compared with the baseline architecture BERT+LSTM, ERASER~\cite{7} improve the model accuracy and AUC by 32.8\% and 36.2\%, and boost the explanation quality by 374.1\%,  -26.2\%, 1290.7\%, and 568.8\% in terms of comprehensiveness, sufficiency, IoU, and explanatory F1 scores, respectively, while FRESH~\cite{31} improves model accuracy, AUC, Sufficiency and Exp F1 by 9.0\%, 10.5\%, 433.3\%,  -26.2\%, 1223.3\%, and 548.2\% respectively. ERASER~\cite{7} has better performance in terms of both model performance as well as explanation quality. Carton et al.~\cite{10}, which employs the BERT+BERT architecture, increases accuracy and AUC by 10.3\%, and 15.5\% and ERASER~\cite{7} achieves the second-best result with an architecture of BERT+LSTM. Expred~\cite{19} obtain the highest explanation correctness and lowest sufficiency with a model architecture of BERT+GRU+MLP, and FRESH~\cite{31}(BERT+LSTM) holds the highest comprehensiveness score among the selected models.

For the MultiRC dataset, Carton et al.~\cite{10} achieves the highest classification accuracy as well as the highest explanation faithfulness and correctness. It improves the model accuracy and AUC by 3.2\%, 15.5\%, lowers sufficiency score by 50.6\%, and boost IoU and explanation F1 by 522.4\%, and 174.4\%, respectively, compared with the baseline. For all models with the architecture BERT+LSTM, while they consistently obtain better results than baseline except for comprehensiveness ERASER~\cite{7}, outperforms FRESH~\cite{31} by 5.3\%, 4.9\%, 2.5\%, and 1.6\% in terms of model accuracy, AUC, IoU, and explanatory F1. FRESH~\cite{31} is more accurate when assessing explanation faithfulness through the sufficiency and comprehensiveness score, with a 14.4\% decrease and 146.2\% boost compared with ERASER~\cite{7}. 

\begin{figure*}
    \centering
    \includegraphics[width=0.9\textwidth]{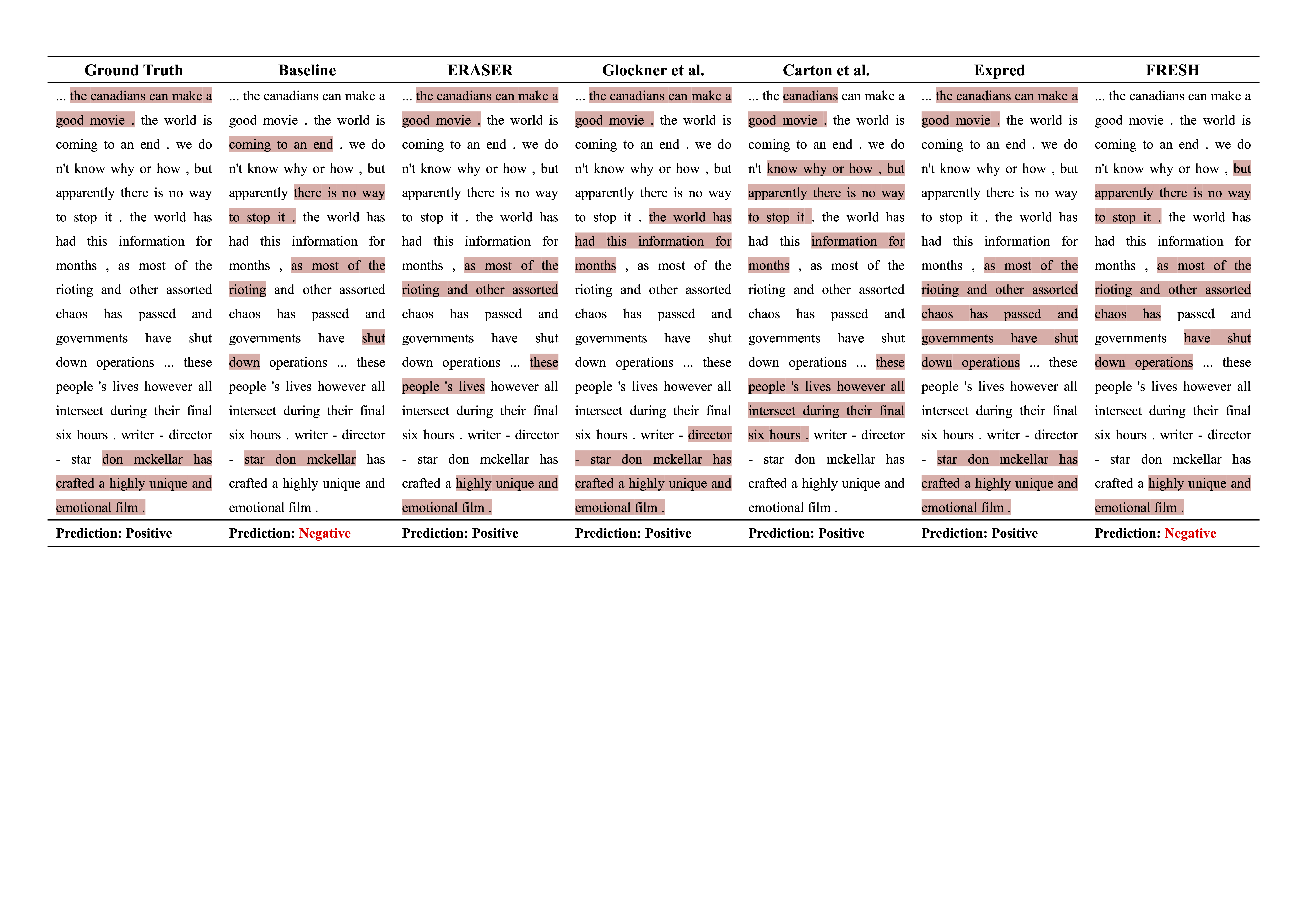}
    \vspace{-100pt}
    \caption{Selected explanation visualization results on FEVER dataset. The model-generated explanations are highlighted.}
    \label{fig:exp_viz1}
\end{figure*}

The performance varies for the FEVER dataset, as FRESH~\cite{31} achieves the highest accuracy, AUC, and sufficiency scores, and Expred~\cite{19} yields the highest comprehensiveness, IoU, Explanatory F1 and Explanatory Recall. All the models perform generally well in the fact verification task in terms of accuracy, with a range of 0.835 to 0.903. In terms of explanatory faithfulness, FRESH~\cite{31} performs worse than the baseline in comprehensiveness but reduces sufficiency by 46.5\%. Expred~\cite{19} obtain the greatest boost in comprehensive and sufficiency, with a change of 394.1\% and -59.0\% respectively. The baseline models generally show poor performance in explanation faithfulness and correctness, which are improved significantly across all five models. Expred~\cite{19} is able to improve IoU by 1863.9\% and Explanatory F1 by 471.0\%. 

\subsubsection{Qualitative Case Study}

\begin{figure*}
    \centering
    \includegraphics[width=0.9\textwidth]{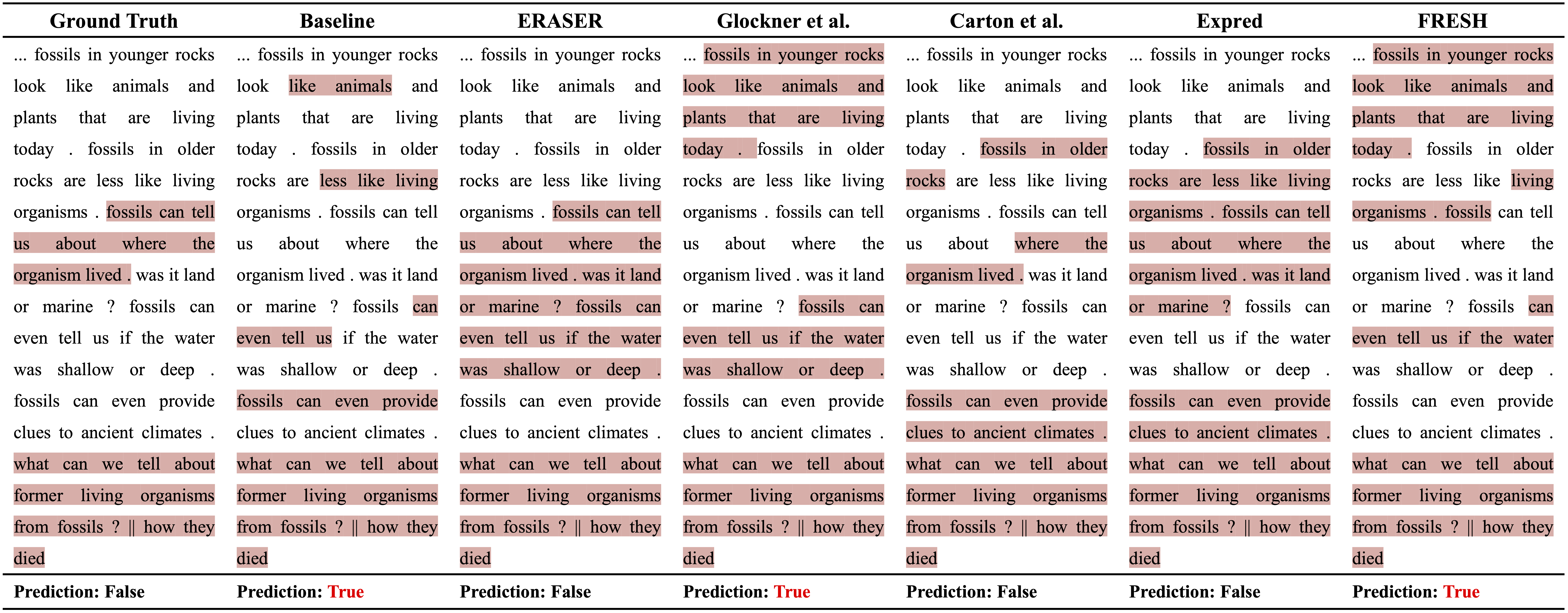}
    \caption{Selected explanation visualization results on Movie Review dataset. The model-generated explanations are highlighted.}
    \label{fig:exp_viz2}
\end{figure*}

\begin{figure*}
    \centering
    \includegraphics[width=0.9\textwidth]{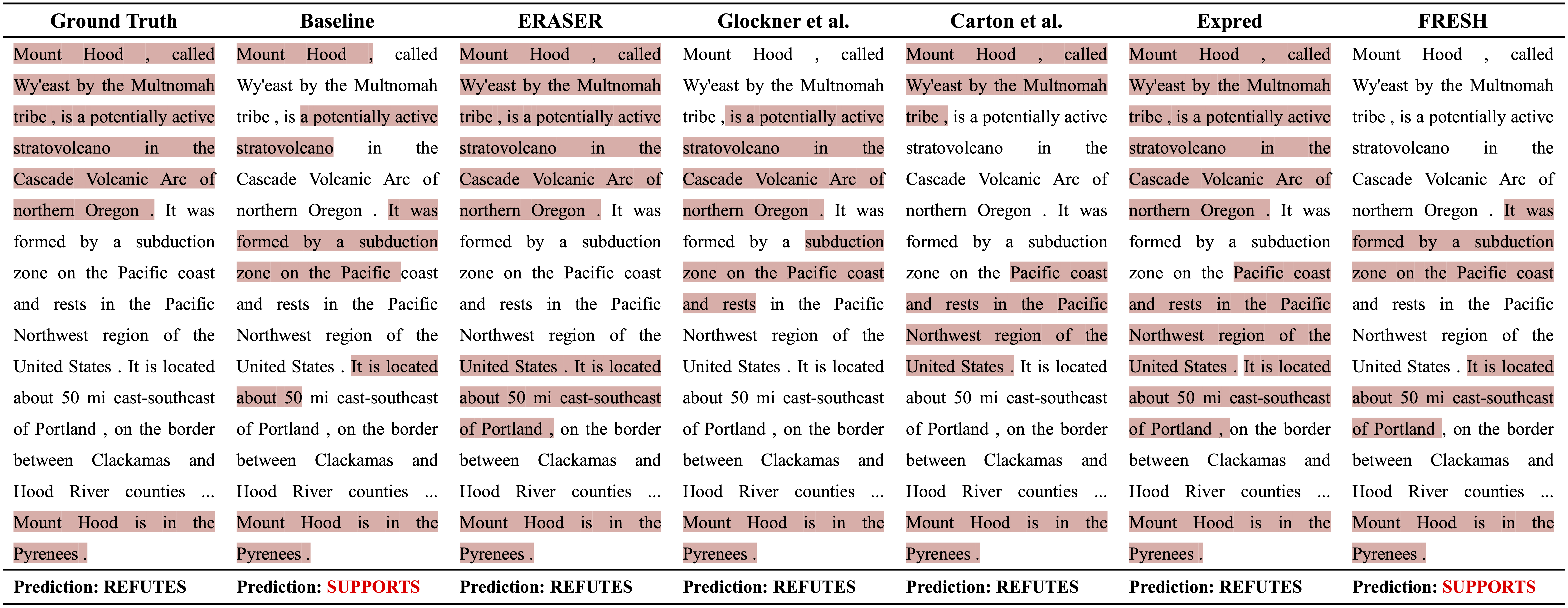}
    \caption{Selected explanation visualization results on MultiRC dataset. The model-generated explanations are highlighted.}
    \label{fig:exp_viz3}
\end{figure*}

Figures \ref{fig:exp_viz1}, \ref{fig:exp_viz2}, and \ref{fig:exp_viz3} provide examples of visualization results on FEVER, Movie Review, and MultiRC dataset. The model-generated explanations are highlighted. In general, the baseline model highlights areas that are scattered all around the corpus, whereas trained models generate explanation rationales that are more aggregated. In Figure \ref{fig:exp_viz1}, ERASER~\cite{7} and Glockner et al.~\cite{9} are highly aligned with ground truth, aligned with their high performance in IoU. While Expred~\cite{19} obtains the highest accuracy and comprehensiveness, its generated-explanation does not align with the ground truth annotations, which implies that the ground truth labels may not be sufficient for the model to learn the prediction. FRESH~\cite{31} generates explanations that are poorly aligned with the ground truth and outputs a wrong prediction label. 

In Figure \ref{fig:exp_viz2}, while Carton et al.~\cite{10} aligns well with the ground truth, it focuses on a higher percent of tokens, which explains why it slightly underperforms in explanatory precision and comprehensiveness but outperforms in sufficiency. There exhibits a compromise between high accuracy and high explanation quality, as Carton et al.~\cite{10} achieves the highest accuracy but lowest comprehensiveness among the selected models. This examples shows how the amount of attention may manipulate the result of explanation faithfulness. If a high amount of tokens are considered important, sufficiency will be close to 0 and comprehensiveness will be relatively high. Therefore, it's necessary to consider both explanation faithfulness and correctness when analyzing the explanation quality. This example also reveals the importance of a case study, to visualize the quantitative results and understand how attention performs in terms of correctness and faithfulness.

\section{Conclusion}
This survey has presented a comprehensive survey of existing methodologies developed in the field of Explanation-Guided Learning (EGL), a group of techniques that applies XAI-driven insights to steer the DNNs' behavior in realizing iterative model revision.
It provides an extensive overview of the EGL challenges, techniques, applications, evaluation procedures, as well as extensive experimental comparison among existing techniques under popular application areas.
It summarizes the findings of the research presented in more than 150 publications on EGL, the majority of which were released in the last five years.
Concretely, in this survey, the formal definition of EGL and its general learning paradigm is first given, along with an overview of the key factors for EGL evaluation, as well as summarization and categorization of existing evaluation procedures and metrics for EGL are provided. 
Based upon the numerous historical and state-of-the-art works discussed in this survey, the article concludes by discussing the current and potential future application areas of EGL, and provides an extensive experimental study that aims at providing the first comprehensive comparative study among existing EGL models in various popular application domains, such as Computer Vision (CV) and Natural Language Processing (NLP) domains.

% possible future challenges:
% 1. Lack of standard evaluation metric for EGL model
% 2. Missing in-depth theatrical analysis of EGL model
% 3. In need of publicly available benchmark datasets with  annotation label data (especially in CV and VQA domains)

% \begin{acks}
% \end{acks}

\bibliographystyle{ACM-Reference-Format}
\bibliography{sections/99_REF}

% \appendix

\end{document}